\documentclass[runningheads]{llncs}

\usepackage{eccv}

\usepackage{eccvabbrv}

\usepackage{graphicx}
\usepackage{booktabs}
\usepackage{amsmath}
\usepackage{amssymb}
\usepackage{subcaption}
\usepackage{multirow}

\usepackage{colortbl} 
\usepackage{arydshln}
\usepackage[dvipsnames]{xcolor}
\usepackage{pifont} 
\usepackage{tikz} 
\definecolor{tabfirst}{rgb}{1, 0.75, 0.7}
\definecolor{tabsecond}{rgb}{1, 0.85, 0.65}
\definecolor{tabthird}{rgb}{1, 0.96, 0.7}
\usepackage{wrapfig}
\usepackage[accsupp]{axessibility}  

\usepackage[pagebackref,breaklinks,colorlinks,citecolor=eccvblue]{hyperref}

\usepackage{orcidlink}

\begin{document}

\title{UniPR-3D: Towards Universal Visual Place Recognition with Visual Geometry Grounded Transformer} 

\titlerunning{UniPR-3D}

\author{Tianchen Deng\textsuperscript{1}*, Xun Chen\textsuperscript{3}*, Ziming Li\textsuperscript{1}, Hongming Shen\textsuperscript{3}, Shuhao Zhai\textsuperscript{4}, Danwei Wang\textsuperscript{3}, Javier Civera\textsuperscript{2}, 
 Hesheng Wang\textsuperscript{1}
}

\authorrunning{T. Deng et al.}

\institute{\textsuperscript{\rm 1} Shanghai Jiao Tong University
\textsuperscript{\rm 2} I3A, University of Zaragoza, Spain
\textsuperscript{\rm 3} Nanyang Technological University
\textsuperscript{\rm 4}University of Macau
}

\maketitle

\begin{abstract}
Visual Place Recognition (VPR) has been traditionally formulated as a single-image retrieval task. Using multiple views offers clear advantages, yet this setting remains relatively underexplored and existing methods often struggle to generalize across diverse environments. In this work, we introduce UniPR-3D, the first VPR architecture that effectively integrates geometry-aware information from multiple views. UniPR-3D builds on a VGGT backbone capable of encoding multi-view 3D representations, which we adapt by designing feature aggregators and fine-tuning them for the place recognition task. To construct our descriptor, we jointly leverage VGGT's 3D tokens, but also intermediate 2D ones. Based on their distinct characteristics, we design dedicated aggregation modules for 2D and 3D features, allowing our descriptor to capture fine-grained texture patterns while also reasoning across viewpoints. To further enhance generalization, we incorporate both single- and multi-frame aggregation schemes, along with a variable-length sequence retrieval strategy. Our experiments show that UniPR-3D sets a new state of the art, outperforming both single- and multi-view baselines and highlighting the effectiveness of geometry-grounded tokens for VPR. Our code and models will be made publicly available on Github.
We will release the code and datasets publicly on \href{https://github.com/dtc111111/UniPR-3D}{https://github.com/dtc111111/UniPR-3D}.

\end{abstract}

\renewcommand{\thefootnote}{} 
\footnotetext{ The first two authors contribute equally to this paper. Corresponding author: Hesheng Wang}

\section{Introduction}
Visual Place Recognition (VPR) is a core problem in robotics and computer vision, targeting the recognition of previously visited locations from visual observations~\cite{lowry2015visual,garg2021your,masone2021survey,zhang2021visual,Schubert2023vpr,milford2025going,Sferrazza_2025_CVPR}.
VPR plays a key role in fundamental tasks such as SLAM~\cite{orbslam3,galvez2012bags, plgslam,mneslam,tosi2026nerfs} and absolute pose estimation~\cite{taira2018inloc,sarlin2019coarse,zhou2024nerfect,liu2026streamvlo}, which underpin a wide range of applications, including autonomous driving, robotic navigation, virtual and augmented reality.

VPR is typically formulated as a single-view retrieval task, in which a query image is used to retrieve an ordered list of the top-$k$ matching candidates from a database. Existing approaches generally rely on neural backbones to extract features, which are then aggregated using various strategies to form global image descriptors. Images are then represented by their aggregated appearance patterns, which are compared at test time via nearest-neighbor search. Backbone architectures have evolved from ResNet-based~\cite{netvlad,gem,patchvlad} to Vision Transformers (ViTs)~\cite{transvpr,r2former,anyloc}. Task-specific fine-tuned foundation models, in particular DINOv2~\cite{dinov2}, achieve nowadays state-of-the-art performance~\cite{salad,izquierdo2024close,berton2025megaloc}. Feature aggregation strategies have also advanced considerably, from the seminal NetVLAD~\cite{netvlad}, inspired by the handcrafted VLAD aggregation~\cite{vlad}, to pooling-based methods such as GeM~\cite{gem}, MLP-based aggregations like MixVPR~\cite{mixvpr}, and approaches leveraging optimal transport~\cite{salad}.

\begin{wrapfigure}{lt}{0.50\textwidth}
  \centering
  \vspace{-0.2cm}
  \includegraphics[width=0.5\textwidth]{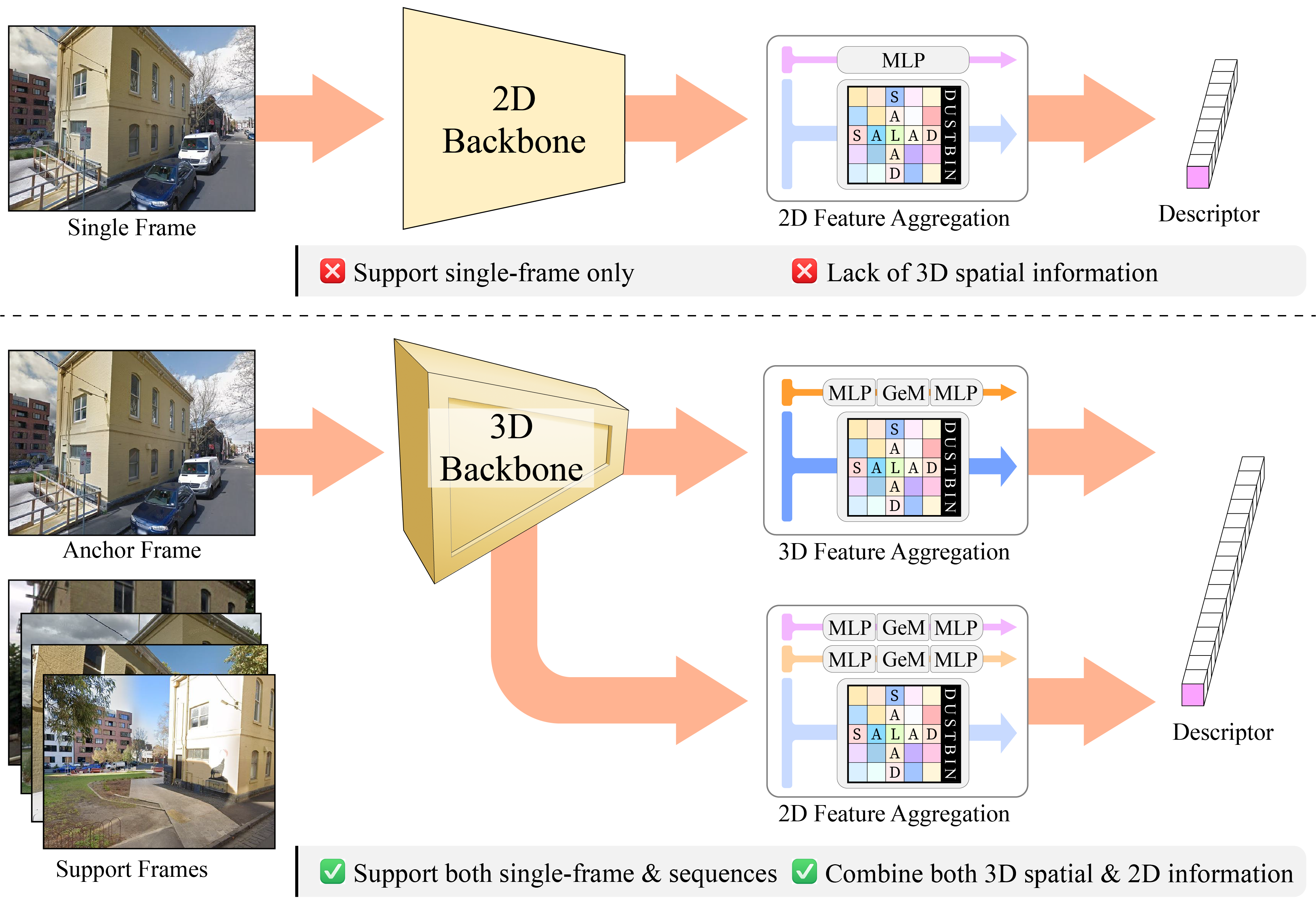}
\vspace{-0.3cm}
  \caption{\textbf{Illustration of a typical single-view VPR model (top) and our multi-view model (bottom).} Single-view VPR extracts image features with a deep backbone and aggregates them into an image descriptor. In contrast, our multi-view VPR model employs a VGGT backbone to jointly extract both 2D and 3D tokens from multiple views, followed by tailored aggregation strategies for each token type. Our framework supports both single-frame and variable-length sequence matching, and achieves state-of-the-art performance across standard VPR benchmarks.  
}
  \vspace{-0.3cm}
  \label{fig:teaser}
\end{wrapfigure}

However, most VPR methods share the fundamental limitation of extracting features from a single view. While multiple views provide broader scene coverage and a richer variety of viewpoints, multi-view feature aggregation has always been a challenging and relatively unexplored topic~\cite{facil2019condition}. The recent development of the Visual Geometry Grounded Transformer (VGGT)~\cite{vggt}, a large ViT backbone encoding 3D-aware multiview representations, has opened new possibilities for leveraging such input modality in various dwonstream tasks~\cite{song2025accelerating,song2026reconvla,liu2026driveva}. Building on this backbone for feature extraction, we introduce in this paper UniPR-3D, the first 3D token-based VPR framework supporting both single-frame and sequence-level place matching.

Our developments work as follows. For single-frame retrieval, we jointly utilize VGGT's 3D tokens and intermediate 2D tokens to construct the place descriptor. Based on the differences between token types, we design distinct aggregation strategies. For 2D tokens, we select the 2D CLS token, 2D register tokens, and 2D patch tokens for aggregation. For the 3D tokens, we discard the camera token (as VPR should be robust to viewpoint changes) and only use the 3D register tokens and 3D patch tokens to ensure stability under viewpoint changes. For patch tokens, we adopt an optimal transport approach~\cite{salad} to derive local patch descriptors, while for register tokens, we employ GeM pooling~\cite{gem} to generate their local descriptors.

For sequence-level retrieval, we define an anchor frame and multiple support frames, and construct a multi-frame feature aggregation method. We use SALAD in order to aggregate both 2D and 3D patch tokens across frames. For 3D register tokens as well as 2D CLS and register tokens, we design an aggregation that combines GeM pooling with an MLP-based projector. The token projector aligns the feature dimensions across different modalities and frames, enabling our network to process input sequences of arbitrary length. This design allows the model to generalize effectively to sequences with varying lengths during inference.

By incorporating 2D and 3D tokens, UniPR-3D achieves universal VPR capabilities across complex environments.
Overall, our contributions are as follows:
\begin{itemize}
    \item We develop the first 3D token-based VPR method with tailored 2D and 3D feature aggregation methods, enabling universal generalization across diverse environments and supporting both single frame and sequence matching.
    \item We propose distinct aggregation strategies for different types of tokens. Specifically, register tokens are GeM-pooled, as their number is relatively small, while patch tokens are aggregated using optimal transport to capture fine-grained spatial correspondences across frames.
    \item We introduce a sequence-level retrieval strategy, with an anchor frame and multiple support frames, and construct a multi-frame feature aggregation method, enabling our network to process input sequences of arbitrary length. 
\end{itemize}

We evaluate UniPR-3D on a comprehensive and diverse set of public datasets under challenging conditions, including temporal variations, spatial changes, and viewpoint shifts. Our results show that our method effectively leverages multi-view information and sets a new state of the art across these benchmarks.

\begin{figure*}[h]
    \centering
    \includegraphics[width=\linewidth]{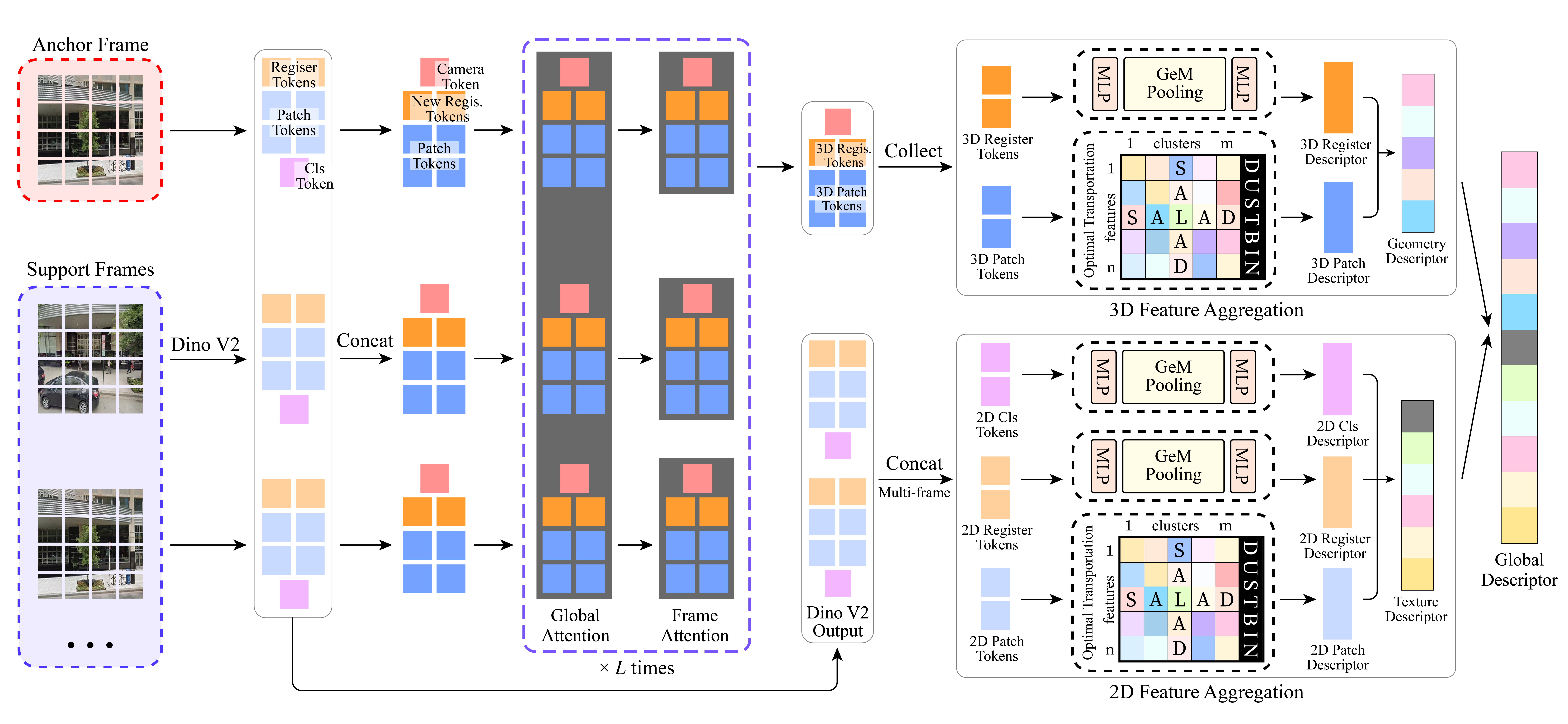}
    \caption{\textbf{Overview.} We propose the first VPR method that supports both single-frame and sequence-level place retrieval. Specifically, we use DINOv2 as our visual feature extractor and then utilize the alternating attention blocks of VGGT to derive 3D tokens. The resulting 3D tokens are divided into different groups, each processed with a dedicated aggregation strategy to form the final descriptor. In particular, cls tokens and register tokens are aggregated using GeM pooling, while patch tokens are processed through an optimal transport module, where the Sinkhorn algorithm is applied to compute the assignment matrix. The outputs from these modules are finally concatenated to produce the global descriptor.}
    \label{fig:system}
\end{figure*}

\section{Related work}
VPR has made remarkable progress in recent years and has become a key component in tasks such as localization~\cite{deng2025mcnslam,grsslam,neslam,mgslam,gong2025dino}, planning and control~\cite{lenav,lv1,lv2,10668846,10486967}, autonomous driving~\cite{prosgnerf,sfpnet,yang1,yang2} and vitual reality~\cite{ma2025controllable,ma2025followcreation,ma2024followpose,chengmoca}. Below, we review the most relevant advances in feature extraction, feature aggregation, and sequence-based retrieval.

\noindent \textbf{Feature Extraction.}
Early VPR methods relied on handcrafted visual descriptors, using either aggregated local features~\cite{vlad} or holistic global representations~\cite{brief,globallocalization}. Geometric and temporal consistency checks were often employed to improve robustness in practical systems~\cite{bow,seqslam}. With the advent of deep learning, ResNet-based architectures became the standard for feature extraction~\cite{netvlad}. More recently, Vision Transformer (ViT)–based foundation models~\cite{3dscenerepresentation} have shown superior representational power and generalization. For example, AnyLoc~\cite{anyloc}, SALAD~\cite{salad}, UniLGL~\cite{unilgl} and so on~\cite{selavpr,selavprpp} leverage DINOv2~\cite{dinov2}: AnyLoc uses it directly, while the rest of them fine-tune it for the VPR task. These models have demonstrated enhanced robustness to variations in illumination, weather, and viewpoint. Nevertheless, the vast majority of existing methods extract features from a single view, thus failing to exploit the spatial information available across multiple observations. The recent development of the 3D foundation model has opened new opportunities for geometry-aware visual representation learning and has shown strong potential in various downstream tasks~\cite{dens3r,MoRE2025,tu2025role,wu2026vega,chen2026out}.

\noindent \textbf{Feature Aggregation.}
Seminal VPR aggregation schemes include Bag of Visual Words (BoW)~\cite{sivic2003video} and Vector of Locally Aggregated Descriptors (VLAD)~\cite{vlad}, both based on clustering handcrafted local descriptors. NetVLAD~\cite{netvlad} extended VLAD by embedding it within a deep architecture that jointly learns the convolutional features and the aggregation process. The Generalized Mean Pooling (GeM)~\cite{gem} of features, proposed later, was a widely used aggregation method combining simplicity and strong retrieval performance. MixVPR~\cite{mixvpr} introduced a MLP-based aggregation module to capture more expressive feature interactions. Recently, SALAD~\cite{salad} proposed an aggregation based on optimal transport~\cite{ot}, achieving the current state of the art in feature aggregation for VPR.

\noindent \textbf{Sequence Retrieval.} 
Most existing approaches~\cite{seqslam,longslam,fast} perform sequence-level retrieval as a post-processing step on top of frame-level similarity scores.
In contrast, several learning-based methods explicitly model temporal dependencies by learning sequential descriptors~\cite{seqmatchnet,seqvlad} or by aggregating consecutive reference and query features into unified sequence-level representations~\cite{seqnet,long-match}. Other works, such as~\cite{sequencematch}, introduce supervised strategies to estimate per-frame sequence-matching confidence, enabling selective fusion of temporal cues. CaseVPR~\cite{casevpr} further proposes a hierarchical sequence-to-frame retrieval pipeline that combines coarse sequence retrieval with fine-grained frame alignment for accurate correspondence estimation. Recent methods like CricaVPR~\cite{lu2024cricavpr} and SeqVLAD~\cite{seqvlad} show promising results. Nevertheless, all of these methods learn to aggregate single-view features without geometric awareness. Despite advances, sequence retrieval methods still exhibit two main limitations.
First, they are typically designed with strong inductive biases, whereas our UniPR-3D, built upon VGGT and our learned aggregation modules, adopts a more data-driven formulation, resulting in greater flexibility.
Second, their aggregation strategies operate purely along the temporal dimension, which makes them sensitive to variations in speed, frame rate, or sequence sparsity.
In contrast, VGGT’s features are geometrically grounded in a 3D-aware multi-view representation, enabling UniPR-3D to achieve greater robustness under such variations.
Differently from the existing VPR literature, our architecture is novel in that it demonstrates for the first time an effective and principled fusion of both 2D single-view and 3D geometry-aware multi-view tokens. This allows our framework to process a variable number of images without strong architectural bias.

\section{Method}
We propose UniPR-3D, a universal visual place recognition framework that supports both frame-to-frame and sequence-to-sequence matching. The input to our method consists of an image sequence $\{\mathbf{I_i}\}$, and for sequence-level matching, we define an anchor frame and multiple support frames. Our model is capable of predicting visual descriptors for both individual frames and entire sequences. The framework is composed of three main modules: (i) Local 3D feature extraction (Sec.~\ref{Sec:extraction});
(ii) Feature assignment and aggregation (Sec.~\ref{Sec:aggregation}). (iii) Sequence matching (Sec.~\ref{Sec:sequencematching}).
We elaborate on the entire pipeline of our system in the following subsections.

\subsection{Local 3D Feature Extraction}
\label{Sec:extraction}

Effective local feature extraction should remain robust under seasonal or day-to-night illumination changes, as well as viewpoint variations. This requires capturing sufficiently stable structural information. However, the most reliable cues for stability lie in the \textbf{3D structural information} of the scene, whereas existing methods primarily focus on texture-level feature extraction from a single image.

Existing models such as AnyLoc~\cite{anyloc} and SALAD~\cite{salad} adopt DINOv2~\cite{dinov2} leverage the exceptional representational capabilities of foundation models to extract local features. However, these approaches can only capture 2D image patterns. In UniPR-3D, we propose the first VPR framework based on the 3D tokens of a VGGT~\cite{vggt} backbone. Our framework follows a ViT-based architecture, in which every input image $\{\mathbf{I_i}\}$ is first divided into patches $m \times m\times c$, which are sequentially processed through transformer blocks to generate different tokens.  

Each image is first processed by the DINOv2 encoder to extract the 2D \textit{cls token}, \textit{register token}, and \textit{patch tokens}, which primarily capture the texture-level characteristics of this image. 
Then we retain only the patch tokens as input to the subsequent alternating attention blocks and initialize two additional tokens: a \textit{camera token}, which encodes the camera extrinsics and intrinsics (including $\{R,T\}$, and the field-of-view parameters), and a \textit{register token}. 
Through the alternating blocks composed of frame attention and global attention, the model produces three types of 3D tokens: a 3D camera token ($n\times1$), 3D register tokens ($n\times4$), and 3D patch tokens ($m\times m$). 
To ensure viewpoint invariance of the final descriptor, we only preserve the 3D register and 3D patch tokens to capture both geometric and structural information of the environment, forming our 3D geometric descriptor.

It is important to note that UniPR-3D does not require any camera intrinsics or extrinsics as input. All descriptors, including the geometry-aware 3D tokens, are computed exclusively from pure RGB images. By leveraging the intermediate VGGT tokens through spatial alternating attention blocks, our representation naturally captures viewpoint-consistent and temporally stable scene structures without relying on external geometric priors.

\subsection{Feature Assignment and Agggregation}
\label{Sec:aggregation}
Due to the heterogeneous nature of 2D and 3D tokens, 
we adopt tailored aggregation strategies for each type. 
In particular, the 2D \textit{cls tokens} ($\mathbf{f}_{cam}$), 2D \textit{register tokens} ($\mathbf{f}_{reg2d}$), and 3D \textit{register tokens} ($\mathbf{f}_{reg3d}$) are limited in number, rendering complex aggregation schemes unnecessary. 
Instead, we employ Generalized Mean (GeM) pooling~\cite{gem}, preceded and followed by lightweight MLPs, 
to generate compact and robust global descriptors:

\begin{equation}
\small
\mathbf{d}_{\text{cls2d}} =
\mathrm{MLP}\!\left(
\left(
\frac{1}{N}\sum_{i=1}^{N}\mathrm{MLP}\left(\mathbf{f}_{cam}^{\,p}\right)
\right)^{\!\!\frac{1}{p}}
\right),
\end{equation}

\begin{equation}
\small
\mathbf{d}_{\text{reg2d}} =
\mathrm{MLP}\!\left(
\left(
\frac{1}{N}\sum_{i=1}^{N}\mathrm{MLP}\left(\mathbf{f}_{reg2d}^{\,p}\right)
\right)^{\!\!\frac{1}{p}}
\right),
\end{equation}

\begin{equation}
\small
\mathbf{d}_{\text{reg3d}} =
\mathrm{MLP}\!\left(
\left(
\frac{1}{N}\sum_{i=1}^{N}\mathrm{MLP}\left(\mathbf{f}_{reg3d}^{\,p}\right)
\right)^{\!\!\frac{1}{p}}
\right),
\end{equation}
where $p$ is a learnable pooling parameter controlling the degree of selectivity.
This simple yet effective aggregation captures the dominant semantic cues of camera and register tokens with high stability.

In contrast, patch tokens are aggregated using an Optimal Transportation (OT) scheme,
which computes a soft matching plan to preserve structural correspondences,
following the design of SALAD~\cite{salad}.
We use two randomly initialized fully connected layers to learn each row of the score matrix $\mathbf{S}$ from scratch for 2D and 3D patch token ${\mathbf{t}}_{2d},{\mathbf{t}}_{3d}$:
\begin{equation}
\mathbf{s}_{2d}=\mathbf{W}_{2d_2}\left(\sigma\left(\mathcal{W}_{2d_1}\left(\mathbf{t}_{2d}\right)+\mathbf{b}_{2d_1}\right)\right)+\mathbf{b}_{2d_2}
\end{equation}
\begin{equation}
\mathbf{s}_{3d}=\mathbf{W}_{3d_2}\left(\sigma\left(\mathcal{W}_{3d_1}\left(\mathbf{t}_{3d}\right)+\mathbf{b}_{3d_1}\right)\right)+\mathbf{b}_{3d_2}
\end{equation}
where $\mathbf{W}_{2d_1},\mathbf{W}_{2d_2},\mathbf{W}_{3d_1},\mathbf{W}_{3d_2}$ are the weights and biases of the layers for 2D tokens, and $\sigma$ is a non-linear activation function. The parameters for the 3D token layers are defined in a similar manner.

Inspired by SALAD~\cite{salad}, we also introduce a dustbin entry for 2D and 3D patch tokens, to which non-informative features can be assigned. We augment the score matrix from $\mathbf{S}$ to $\overline{\mathbf{S}}=\left[\mathbf{S}, \overline{\mathbf{s}}_{i, m+1}\right] \in \mathbb{R}_{>0}^{n \times m+1}$, by appending the column $\overline{\mathbf{s}}_{i, m+1}$ representing the feature-to-dustbin relation. Similar to SuperGlue~\cite{superglue}, this score is modeled with a single learnable parameter $z\in \mathbb{R}$.
\begin{equation}
\overline{\mathbf{s}}_{i, m+1}=z \mathbf{1}_n
\end{equation}
where $\mathbf{1}_n=[1, \ldots, 1]^{\top} \in \mathbb{R}^n$ a n-dimensional vector of ones.
We follow SuperGlue~\cite{superglue} and SALAD~\cite{salad}, and use the Sinkhorn Algorithm~\cite{sinkhorn} to obtain
the optimal assignment $\overline{\mathbf{P}} \in \mathbb{R}^{n \times(m+1)}$:
\begin{equation}
\overline{\mathbf{P}} \mathbf{1}_{m+1}=\boldsymbol{\mu} \quad \text { and } \quad \overline{\mathbf{P}}^{\top} \mathbf{1}_n=\boldsymbol{\kappa} .
\end{equation}
The Sinkhorn Algorithm finds the optimal transport assignment between the distributions $\mu$ and $\boldsymbol{\kappa}$ by iteratively normalizing the rows and columns of the score matrix. Finally, we drop the dustbin column to obtain the assignment $\mathbf{P}=\left[\mathbf{p}_{*, 1}, \ldots, \mathbf{p}_{*, m}\right]$. We use a MLP to reduce the dimensionality of the patch tokens for efficiency.

Then the patch descriptor can be computed as: $d_{j, k}=\sum_{i=1}^n P_{i, k} \cdot f_{i, k}$, where $f_{i, k}$ corresponds to the $k^{\text {th }}$ dimension of $\mathbf{f}_i$, with $k \in \{1, \ldots, l\}$.
The final descriptor is composed of the 2D CLS descriptor, 2D register descriptor, 2D patch descriptor, 3D register descriptor and 3D patch descriptor. These five components are concatenated to form the final unified descriptor.
\begin{equation}
    \mathbf{d}=\left[\mathbf{d}_{cls2d}^\top \ \ \mathbf{d}_{reg2d}^\top \ \ \mathbf{d}_{patch2d}^\top \ \ \mathbf{d}_{reg3d}^\top \ \ \mathbf{d}_{patch3d}^\top\right]^\top
\end{equation}

\subsection{Sequence Matching}
\label{Sec:sequencematching}
For frame-to-frame matching, we directly process a single image to extract its corresponding 3D tokens. For sequence-to-sequence matching, we instead define an anchor frame together with multiple support frames. In the VGGT architecture, the first frame defines the world coordinate system, and all tokens are registered with respect to this anchor. This design, coupled with the training pipeline, makes the first frame essential for preserving spatial consistency across the sequence. 

Our multi-frame aggregation strategy is as follows. 
Existing multi-frame fusion approaches typically operate on fixed-length sequences, 
requiring identical sequence lengths during training and inference, 
which limits their generalization. 
To overcome this constraint, we propose a variable-length sequence fusion framework. Specifically, we design a multi-frame projector based on Generalized Mean (GeM) pooling to aggregate the 2D CLS tokens, 2D register tokens, and 3D register tokens 
across frames, enabling flexible sequence-level feature fusion.
\begin{equation}
\mathbf{d}_{\text{reg3d}}^M =
\mathrm{MLP}\!\left(
\left(
\frac{1}{N}\sum_{i=1}^{N}\mathrm{MLP}\left(\{\mathbf{f}_{reg3d}^{\,p},\mathbf{f}_{reg3d}^{\,p},\dots,\mathbf{f}_{reg3d}^{\,p}\}\right)
\right)^{\!\!\frac{1}{p}}
\right),
\end{equation}
For the multi-frame patch tokens, we adopt an optimal transport 
formulation for feature aggregation. 
Unlike the single-frame case, we first cluster the patch tokens 
belonging to different frames and then compute the assignment 
matrix using the Sinkhorn algorithm. 
The final patch descriptor is computed as:
\begin{equation}
    \mathbf{d}_{patch2d}^M= \sum_{i=1}^n P_{i, k}^M \cdot \{f_{i, k}^1,f_{i, k}^2,\dots,f_{i, k}^M\}
\end{equation}
\subsection{Training}
We follow the experimental setups of two recent works: 
SALAD~\cite{salad} for single-frame retrieval 
and SeqMatchNet~\cite{seqmatchnet} for sequence-level retrieval.

For single-frame training, we use GSV-Cities~\cite{gsv} as it is standard practice in the field. GSV-Cities is a large-scale collection of urban locations from Google Street View. For optimization, we adopt the multi-similarity loss~\cite{multiloss} together with the AdamW optimizer~\cite{adam}. For sequence-level training, since GSV-Cities 
does not support multi-frame training, we follow the setting of 
SeqMatchNet~\cite{seqmatchnet} and train our model on the 
Mapillary Street-Level Sequences (MSLS) dataset~\cite{msls}. 
MSLS contains over 1.6 million images collected from the Mapillary 
collaborative mapping platform, covering diverse cities and exhibiting 
significant variations in viewpoint, weather, and illumination conditions.
The training process consists of two stages: in the first stage, we train only the descriptor head; 
in the second stage, we jointly train the alternating attention blocks 
of VGGT and the DINOv2 encoder. To ensure a fair and accurate comparison, especially with strong foundation model-based baselines like SALAD~\cite{salad}, we train our models using full-precision (float32). Furthermore, we align the input image resolution across models, which not only standardizes the evaluation but also optimizes the inference latency of our framework.

\section{Experiments}
To rigorously evaluate the effectiveness of our proposed method, we conduct extensive experiments following the standard evaluation protocols established in SALAD~\cite{salad} for single-frame retrieval and follow SeqMatchNet~\cite{seqmatchnet} for sequence retrieval. 

\subsection{Implementation Details}
Our model architecture follows VGGT~\cite{vggt} with $L = 24$ alternating frame and global attention layers. To accelerate inference, we incorporate FlashAttention-2~\cite{flashattention}. The model is initialized using pre-trained VGGT weights, and is trained for VPR using the AdamW optimizer with a hybrid learning rate schedule: a linear warm-up over the first 0.5 epochs, followed by cosine decay, with a peak learning rate of $1\times 10^{-6}$. All training are conducted on a NVIDIA A100 GPU. The inference speed is tested on a single NVIDIA RTX 4090 GPU. For the 3D backbone, we adopt a LoRA-based fine-tuning strategy, where both the frame attention and global attention blocks are refined. For the cls descriptor and register descriptor, we employ dimensionality reduction, compressing the feature token dimensions to 256. In contrast, the patch descriptor is represented with a dimension of 
$128\times 64=8192$, with 64 clusters. 

\noindent \textbf{Dataset and Metrics}
For single-frame evaluation,
we conduct experiments on a wide range of benchmarks to verify the generalization of our method. We use the validation and challenge partitions of MSLS~\cite{msls}, which consists of dashcam images, Pittsburgh250k-test~\cite{Pittsburg}, which images urban scenarios, NordLand~\cite{nordland} for its seasonal variations in train-front imagery in Norway, SPED~\cite{sped} which records with surveillance cameras, Tokyo 24/7~\cite{tokyo247} including differences in light between day and night, and SF-XL~\cite{sfxl,sfxl_extraqueries}, containing heavy illumination changes and substantial occlusions. We use Recall@k (R@k) as the evaluation metric for all experiments, following standard practice in the literature. We follow the MixVPR evaluation protocol~\cite{mixvpr}, where retrieval is considered correct if at least one image within 25 meters of the query location (or within two frames for the NordLand dataset) appears among the top-k candidates. 

For multi-frame evaluation, we conduct experiments on the 
Nordland~\cite{nordland}, MSLS~\cite{msls}, Oxford1~\cite{oxfordrobo}, and Oxford2~\cite{oxfordrobo} datasets. Specifically, for Oxford1 we use 2014-12-16-18-44-24 (winter night) to 2014-11-
18-13-20-12 (fall day), sampled every 2 meters. And for Oxford2, 2014-11-14-16-34-33 (fall night) to 2015-11-13-
10-28-08 (fall day), sampled every 2 meters.
We adopt Recall@k (R@k) as the evaluation metric for all experiments. 
Differently from the single-frame setting, for the Oxford datasets, 
directly using a retrieval threshold of 25 meters often leads to 
saturated results (close to 100\%) for existing methods. 
Therefore, we perform a more detailed evaluation using thresholds 
of 2\,m and 25\,m. 
For the Nordland dataset, retrieval is considered correct if the 
ground-truth frame within ten frames from the query appears 
among the top-\emph{k} retrieved candidates. Following CaseVPR~\cite{casevpr}, the sequence length S is set to be 5.

\begin{table*}[h]
\centering
\resizebox{\linewidth}{!}{
\setlength{\tabcolsep}{1mm}{
    \begin{tabular}{lcccccccccccccccccccc}
\toprule
\multirow{2}{*}{Method} & \multicolumn{2}{c}{MSLS Chall.} & \multicolumn{2}{c}{MSLS val} & \multicolumn{2}{c}{Pitts250k} & \multicolumn{2}{c}{Nordland} & \multicolumn{2}{c}{SPED} & \multicolumn{2}{c}{SF-XL v1} & \multicolumn{2}{c}{SF-XL v2} & \multicolumn{2}{c}{SF-XL ni.} & \multicolumn{2}{c}{SF-XL occ.} & \multicolumn{2}{c}{Tokyo 24/7} \\
\cmidrule(lr){2-3} \cmidrule(lr){4-5} \cmidrule(lr){6-7} \cmidrule(lr){8-9} \cmidrule(lr){10-11} \cmidrule(lr){12-13} \cmidrule(lr){14-15} \cmidrule(lr){16-17} \cmidrule(lr){18-19} \cmidrule(lr){20-21}
  & R@1 & R@10 & R@1 & R@10 & R@1 & R@10 & R@1 & R@10 & R@1 & R@10 & R@1 & R@10 & R@1 & R@10 & R@1 & R@10 & R@1 & R@10 & R@1 & R@10 \\
\midrule
NetVLAD\cite{netvlad} & 35.1 & 51.7 & 54.5 & 70.4 & 85.9 & 95.0 & 32.6 & 53.3 & 78.7 & 91.4 & 40.1 & 57.7 & 76.9 & 91.1 & 6.7 & 14.2 & 9.2 & 22.4 & 69.8 & 82.9 \\
AP-GeM\cite{apgem} & 35.9 & 53.6 & 56.0 & 72.9 & 80.0 & 93.5 & 22.9 & 42.1 & 64.6 & 83.5 & 37.9 & 54.1 & 66.4 & 84.6 & 7.5 & 16.7 & 5.3 & 14.5 & 57.5 & 77.5 \\
CosPlace\cite{cosplace} & 67.2 & 76.4 & 85.0 & 92.6 & 92.3 & 98.4 & 44.2 & 54.4 & 75.9 & 89.1 & 76.6 & 85.5 & 88.8 & 96.8 & 23.6 & 32.8 & 30.3 & 44.7 & 87.3 & 95.6 \\
MixVPR\cite{mixvpr} & 64.0 & 80.7 & 83.2 & 91.9 & 94.3 & 98.9 & 58.4 & 80.0 & 85.2 & 94.6 & 72.5 & 80.9 & 88.6 & 95.0 & 19.5 & 30.5 & 30.3 & 38.2 & 87.0 & 94.0 \\
EigenPlaces\cite{eigenplaces} & 67.4 & 81.7 & 85.9 & 93.1 & 94.1 & 98.7 & 54.2 & 73.9 & 82.4 & 94.7 & 84.0 & 90.7 & 90.8 & 96.7 & 23.6 & 34.5 & 32.9 & 52.6 & 93.0 & 97.5 \\
AnyLoc\cite{anyloc} & 40.2 & 51.3 & 58.7 & 74.5 & 89.4 & 98.0 & 69.5 & 84.9 & 80.5 & 92.5 & 82.7 & 88.5 & 88.9 & 94.8 & 20.5 & 31.9 & 29.3 & 39.9 & 87.6 & 97.5 \\
SALAD\cite{salad} & 75.0 & 91.3 & 92.2 & 97.2 & 95.2 & 99.2 & 76.0 & 90.5 & 92.1 & 97.2 & 90.9 & 95.9 & 95.1 & 98.7 & 50.3 & 68.4 & 51.8 & 74.1 & 95.1 & 98.3 \\
CricaVPR\cite{lu2024cricavpr} & 69.3 & 85.9 & 76.7 & 87.2 & 92.6 & 98.3 & 70.2 & 85.5 & 83.1 & 93.5 & 62.6 & 78.9 & 86.3 & 96.0 & 25.8 & 40.6 & 27.6 & 47.4 & 82.9 & 93.7 \\
BoQ\cite{ali2024boq} & 71.2 & 86.9 & 91.2 & 96.1 & 95.0 & 99.1 & 70.7 & 87.5 & 86.5 & 95.7 & 87.9 & 93.7 & 88.5 & 94.9 & 27.4 & 37.2 & 34.7 & 54.1 & 97.5 & 98.7 \\
MegaLoc\cite{berton2025megaloc} & 73.4 & 88.7 & 91.0 & 95.8 & 96.4 & 99.3 & 76.7 & 90.9 & 92.0 & 95.9 & \underline{95.3} & \underline{98.0} & 94.8 & 98.5 & \underline{52.8} & \underline{73.8} & 51.3 & 75.0 & 96.5 & 99.4 \\
\midrule
UniPR 3D  & \underline{75.5} & \underline{91.4} & \underline{92.9} & \underline{97.6} & \underline{96.5} & \underline{99.5} & \underline{78.4} & \underline{91.4} & \underline{92.6} & \underline{97.7} & 93.8 & 96.9 & \underline{95.7} & \underline{99.1} & 52.1 & 72.8 & \underline{52.1} & \underline{75.8} & \underline{97.6} & \underline{99.5} \\
UniPR 3D*  & \textbf{75.9} & \textbf{91.8} & \textbf{93.2} & \textbf{97.9} & \textbf{96.6} & \textbf{99.7} & \textbf{78.9} & \textbf{92.4} & \textbf{92.8} & \textbf{97.9} & \textbf{96.3} & \textbf{98.5} & \textbf{95.9} & \textbf{99.3} & \textbf{54.8} & \textbf{74.5} & \textbf{53.9} & \textbf{77.4} & \textbf{97.9} & \textbf{99.7}  \\
\bottomrule
\end{tabular}
}}
    \caption{\textbf{Single-frame matching results.} We compare our UniPR-3D against existing single-frame VPR baselines. Our method achives a noticeable improvement in the recall, highlighting the benefit of combining 2D and 3D tokens for the place retrieval. * indicates same training setting as MegaLoc.}
    \label{tab:singleframe}
\end{table*}

\begin{table*}[h]
\centering
\resizebox{\linewidth}{!}{
\setlength{\tabcolsep}{1.1mm}{
\begin{tabular}{lcccccccccccccc}
\toprule
\multirow{2}{*}{Method} & \multirow{2}{*}{Backbone} & \multirow{2}{*}{Desc. size} & \multicolumn{3}{c}{MSLS Val (pos=25m)} & \multicolumn{3}{c}{NordLand (pos=10f)} & \multicolumn{3}{c}{Oxford1 (pos=2m)} & \multicolumn{3}{c}{Oxford2 (pos=2m)} \\
\cmidrule(lr){4-6} \cmidrule(lr){7-9} \cmidrule(lr){10-12} \cmidrule{13-15}
 & & & R@1 & R@5 & R@10 & R@1 & R@5 & R@10 & R@1 & R@5 & R@10 & R@1 & R@5 & R@10 \\
\midrule
SeqSLAM~\cite{seqslam} & VGG-16 & 4096 & 45.9 & 58.2 & 70.4 & 53.1 & 67.4 & 71.9 & 34.7 & 51.1 & 70.2 & 26.5 & 36.9 & 44.0\\
SeqMatchNet~\cite{seqmatchnet} & VGG-16 & 4096 & 65.5 & 77.5 & 80.3 & 56.1 & 71.4 & 76.9 & 36.8 & 43.3 & 48.3 & 27.9 & 38.5 & 45.3\\
Delta Descriptors~\cite{delta} & VGG-16 & 4096 & 43.0 & 58.4 & 68.6 & 53.1 & 67.4 & 70.2 & 36.9 & 55.2 & 74.1 & 26.5 & 46.3 & 62.8\\
NetVLAD + FC~\cite{netvlad} & ResNet-18 & 4096 & 68.5 & 79.1 & 84.6 & 59.3 & 75.5 & 80.0 & 50.1 & 60.4 & 71.3 & 18.1 & 28.5 & 35.2 \\
NetVLAD + FC~\cite{netvlad} & ResNet-50 & 4096 & 71.0 & 80.5 & 84.8 & 64.6 & 67.2 & 78.4 & 58.1 & 72.4 & 78.3 & 20.1 & 31.5 & 39.2 \\
SeqPool + CAT & CCT224 & 384$\cdot$SL & 74.2 & 83.5 & 87.9 & 62.6 & 65.2 & 76.4 & 52.5 & 65.4 & 73.3 & 13.9 & 26.4 & 33.2\\
SeqPool + CAT & CCT384 & 384$\cdot$SL & 77.8 & 85.0 & 88.1 & 63.5 & 65.9 & 76.6 & 53.5 & 66.6 & 74.3 & 14.5 & 27.4 & 34.2\\
GeM + CAT~\cite{gem} & ResNet-18 & 1280 & 76.8 & 84.4 & 89.8 & 61.2 & 63.1 & 74.7 & 57.5 & 71.6 & 79.3 & 19.5 & 30.4 & 38.2\\
GeM + CAT & ResNet-50 & 5120 & 68.6 & 78.9 & 84.7 & 62.2 & 64.1 & 75.9 & 59.6 & 73.1 & 81.4 & 20.4 & 30.4 & 38.1\\
SeqNet~\cite{seqnet} & ResNet-18 & 4096 & 68.0 & 77.6 & 82.5 & 60.4 & 62.5 & 73.1 & 52.4 & 65.8 & 74.8 & 14.1 & 25.9 & 33.3\\
SeqNet~\cite{seqnet} & ResNet-50 & 4096 & 71.1 & 80.2 & 85.0 & 61.9 & 64.4 & 75.8 & 57.4 & 69.8 & 76.8 & 16.5 & 27.9 & 36.3\\
JIST~\cite{jist} & ResNet-18 & 512 & 86.6 & 89.4 & 91.5 & 62.3 & 72.4 & 88.1 & 57.2 & 69.8 & 77.8 & 17.1 & 27.9 & 35.3 \\
sVPR~\cite{svpr} & ResNet-50 & 24576 & 87.6 & 90.4 & 92.1 & 63.5 & 73.9 & 88.9 & 58.3 & 70.4 & 78.4 & 17.8 & 28.4 & 36.5 \\
SeqVLAD~\cite{seqvlad} & CCT384 & 24576 & 89.9 & 92.4 & 94.1 & 65.5 & 75.2 & 80.0 & 58.4 & 72.8 & 80.8 & 19.1 & 29.9 & 37.3 \\
CaseVPR~\cite{casevpr} & DinoV2 Vit-B & 10752 & \underline{91.2} & \underline{94.1} & \underline{95.0} & \underline{84.1} & \underline{89.9} & \underline{92.2} & \underline{90.5} & \underline{95.2} & \underline{96.5} & \underline{72.8} & \underline{85.8} & \underline{89.9}\\
\midrule
\textbf{UniPR-3D} (ours) & VGGT & 17152 & \textbf{93.7} & \textbf{95.7} & \textbf{96.9} & \textbf{86.8} & \textbf{91.7} & \textbf{93.8} & \textbf{95.4} & \textbf{98.1} & \textbf{98.7} & \textbf{80.6} & \textbf{90.3} & \textbf{93.9} \\
\bottomrule
\end{tabular}
}}
\caption{\textbf{Sequence matching results.} 
We compare our UniPR-3D against existing sequence-level VPR baselines. Our method achieves significantly higher recall that competing approaches, highlighting the advantages of our 
3D token–based framework in aggregating image patterns 
across multiple views. Note that for the Oxford datasets, the retrieval distance is set to 2\,m.}
\label{tab:sequence}
\vspace{-0.4cm}
\end{table*}

\begin{table}[h]
\centering
\scalebox{0.78}{
\setlength{\tabcolsep}{1mm}{
\begin{tabular}{lccccccc}
\toprule
\multirow{2}{*}{Methods} &\multirow{2}{*}{Latency} & \multicolumn{3}{c}{Oxford1 (pos=2m)}   &\multicolumn{3}{c}{Oxford1 (pos=25m)}\\
\cmidrule(lr){3-5} \cmidrule(lr){6-8} 
& & R@1 & R@5 & R@10 & R@1 & R@5 & R@10 \\
\midrule
SeqMatchNet~\cite{seqmatchnet} & 115 & 36.8  &  43.3  &  48.3  & 47.2  & 54.7 & 63.4  \\
SeqVLAD~\cite{seqvlad} & 129 & 58.4 & 72.8 & 80.8 & 72.2 & 80.3 & 86.1  \\
CaseVPR~\cite{casevpr} & 75 & 90.5 & 95.2 & 96.5 & 97.8 & 98.8 & 99.3  \\
\midrule
\textbf{UniPR-3D} (ours) & 140 & \textbf{95.4}  & \textbf{98.1} & \textbf{98.7} & \textbf{99.3} & \textbf{99.5} & \textbf{99.6}  \\
\bottomrule
\end{tabular}}}
\caption{\textbf{Comparison against baselines under varying distance thresholds 
(2\,m and 25\,m) in the Oxford dataset~\cite{oxfordrobo}.} Our method consistently outperforms all baselines 
across all settings.\\[-0.8cm]}
\label{tab:seqence2}
\end{table}

\begin{figure}[h]
\centering
    \includegraphics[width=0.9\linewidth]{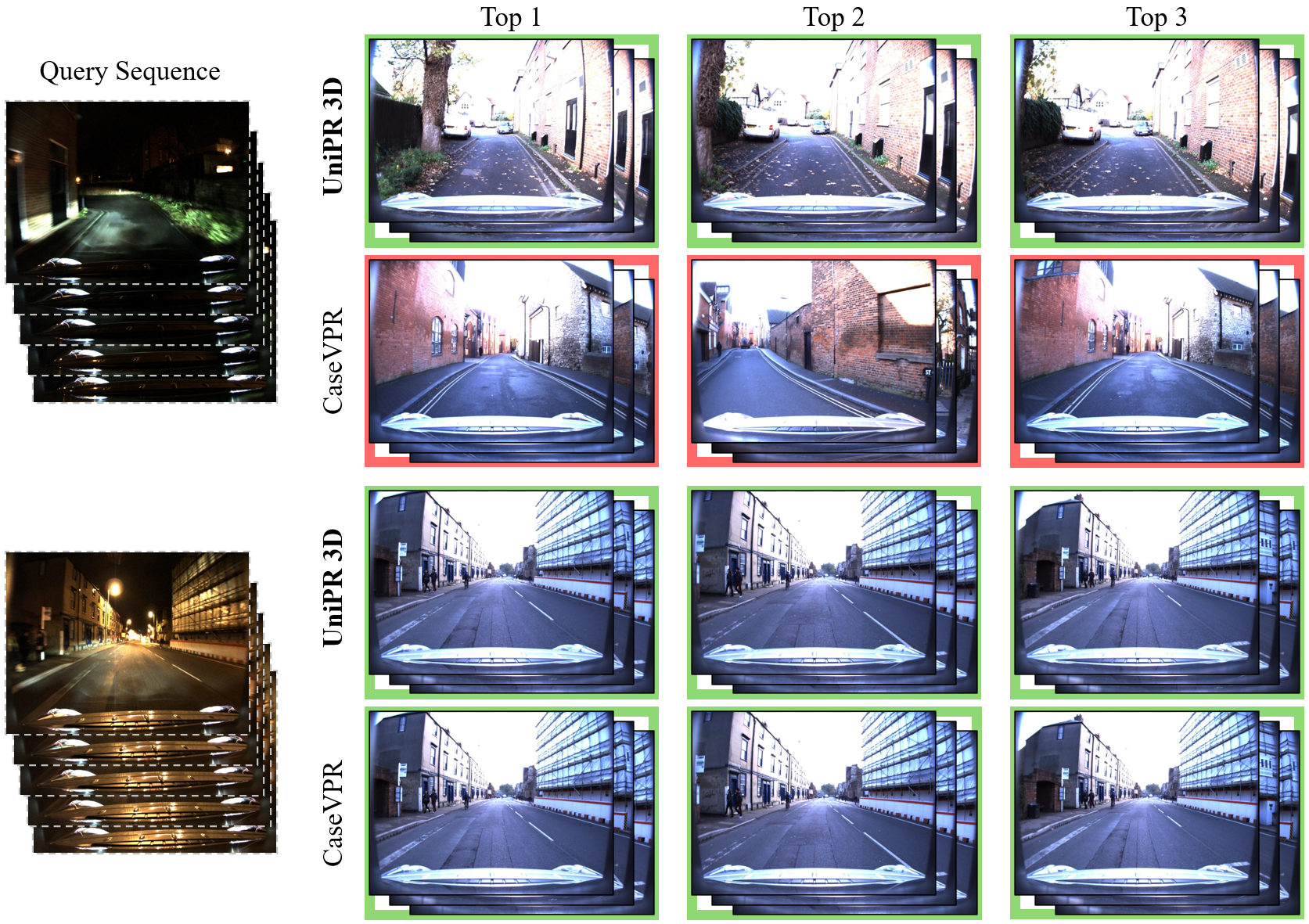}
    \caption{\textbf{Qualitative sequence matching results} on the Oxford dataset~\cite{oxfordrobo}.
The left column shows two query sequences, while the three right columns present the top-3 candidates 
retrieved by our UniPR-3D and the baseline CaseVPR~\cite{casevpr}. 
Successful retrievals are framed in \textcolor{green}{green}, 
while erroneous ones are shown in \textcolor{red}{red}.  
UniPR-3D retrieves the correct place even under seasonal, weather, viewpoint, 
and day-night variations.\\[-0.8cm]}
    \label{fig:topk}
    \vspace{-0.4cm}
\end{figure}

\begin{figure}[h]
\centering
    \includegraphics[width=0.8\linewidth]{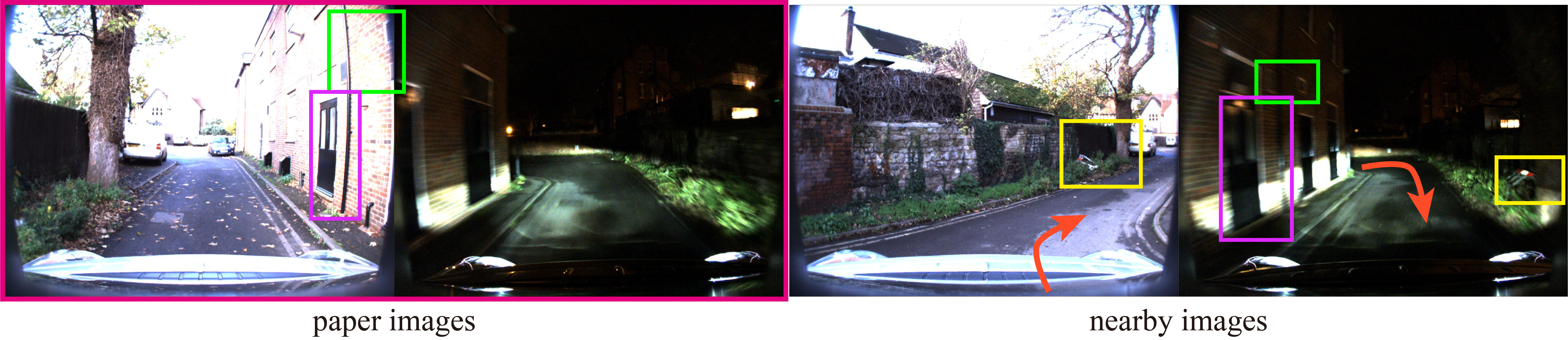}
    \vspace{-0.4cm}
    \caption{An illustration explicitly showing the viewpoint-reversed behavior.
The two sequences involved are captured under daytime and nighttime conditions, respectively, with the vehicle traversing the same route in opposite directions. As a result, the retrieved images naturally exhibit viewpoint reversal with respect to the queries, leading to front–back flipped perspectives in the visualization.}
    \label{fig:flip_comparison}
    \vspace{-0.4cm}
\end{figure}


\subsection{Experimental Results}

\noindent\textbf{Single-Frame Matching.} We compare against a diverse set of baselines, including NetVLAD~\cite{netvlad}, AP-GeM~\cite{apgem}, CosPlace~\cite{cosplace}, MixVPR~\cite{mixvpr}, EigenPlaces~\cite{eigenplaces}, AnyLoc~\cite{anyloc}, SALAD~\cite{salad}, CricaVPR~\cite{lu2024cricavpr}, BoQ~\cite{ali2024boq}, and MegaLoc~\cite{berton2025megaloc}. As shown in Tab.~\ref{tab:singleframe}, UniPR-3D outperforms all baselines on most datasets 
and across most metrics. 
All compared methods are trained under the same setting using the 
GSV-Cities dataset, and the results of the baseline models are obtained 
by retraining the official implementations released by the authors. We further evaluate our approach on challenging datasets, 
including MSLS Challenge, characterized by its large scale, diversity, 
and closed-set labeling, and Nordland, known for its extreme 
visual similarity and severe seasonal variations. 
The experimental results demonstrate that our 3D token–based descriptor 
is capable of capturing stable 3D structural information within the scene, 
rather than relying solely on texture cues, leading to significantly improved 
robustness and generalization. 
As expected, our method incurs in higher latency compared 
to existing approaches due to the incorporation of additional 3D information.

\noindent\textbf{Sequence Matching.} We compare against a diverse set of existing sequence-matching VPR methods: SeqSLAM~\cite{seqslam},  SeqNet~\cite{seqnet}, SeqMatchNet~\cite{seqmatchnet}, Delta Descriptors~\cite{delta}, JIST~\cite{jist}, sVPR~\cite{svpr}, SeqVLAD~\cite{seqvlad} and CaseVPR~\cite{casevpr}. We retrain all competing models on the MSLS dataset to ensure a fair comparison. 
As shown in Tab.~\ref{tab:sequence}, our method delivers superior performance 
on multi-frame matching tasks, consistently outperforming prior approaches 
across all datasets and evaluation metrics. 
In Tab.~\ref{tab:seqence2}, we additionally report results on the Oxford dataset 
under different retrieval distance thresholds (2\,m, 25\,m). 
We find that the stricter 2\,m threshold more clearly exposes performance differences among methods. 
Under this setting, UniPR-3D surpasses the previous state of the art by more than \textbf{10}\%. 
These results clearly show the effectiveness of UniPR-3D 
in capturing 3D structural cues from multiple views, and maintaining robustness 
against variations in illumination, weather, and viewpoint. 
By leveraging spatial information across frames in a geometry-aware manner, our sequence-level framework highlights the significant benefits and future potential of incorporating 3D tokens into VPR systems.

\begin{figure*}[h]
    \centering
    \includegraphics[width=0.95\linewidth]{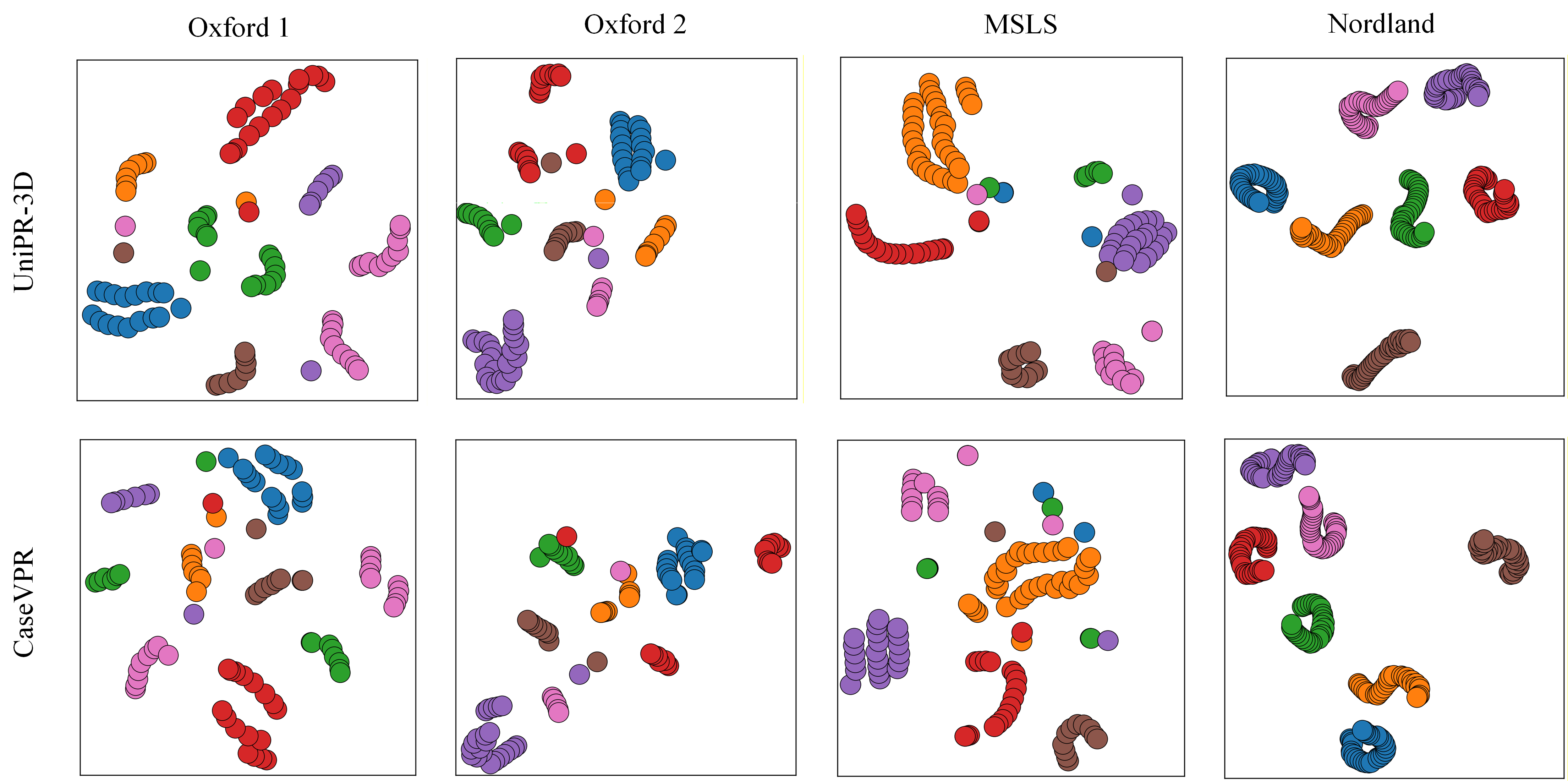}

    \caption{\textbf{t-SNE visualization} comparing our UniPR-3D descriptor against CaseVPR across different datasets. }
    \label{fig:tsne}
\vspace{-0.4cm}
\end{figure*}

Besides the quantitative results, we also visualize several sequence matching examples in Fig.~\ref{fig:topk}, where we show the query frames together with their top-3 sequences retrieved by UniPR-3D. We further compare our method with the current SOTA sequence matching method CaseVPR~\cite{casevpr}. UniPR-3D is able to retrieve correct matches even under challenging conditions, such as severe illumination changes and large viewpoint variations, further confirming the robustness of our 3D token–based descriptor.
As visualized in Fig.~\ref{fig:topk}, some retrieved images naturally exhibit a flipped front-back perspective with respect to queries, as shown in Fig.~\ref{fig:flip_comparison}. This occurs because the corresponding sequences in the Oxford dataset are captured under daytime and nighttime conditions with the vehicle traversing the same route in opposite directions. Importantly, this viewpoint-reversed behavior does not indicate incorrect retrieval; on the contrary, it strongly reflects the robust place recognition capabilities of our 3D geometry-grounded descriptor under extreme viewpoint and illumination changes.
As shown in Fig.~\ref{fig:tsne}, we provide t-SNE visualization. Our method produces well-separated and compact clusters, indicating strong discriminative capability and consistent feature structuring. This qualitative comparison further demonstrates the superiority of our 3D token–enhanced representation in capturing robust geometric and semantic information for visual place recognition.

\begin{table*}[h]
\centering
\resizebox{\linewidth}{!}{
\setlength{\tabcolsep}{1mm}{
\begin{tabular}{ccccccccccccccc}
\toprule
\multirow{2}{*}{2D cls} & \multirow{2}{*}{2D regis.} & \multirow{2}{*}{2D patch} & \multirow{2}{*}{3D regis.} & \multirow{2}{*}{3D patch OT}
& \multirow{2}{*}{3D patch GEM} & \multirow{2}{*}{3D Pose}& \multicolumn{2}{c}{MSLS val} & \multicolumn{2}{c}{Oxford 1} & \multicolumn{2}{c}{Oxford 2}
& \multicolumn{2}{c}{Nordland}\\
\cmidrule(lr){8-9} \cmidrule(lr){10-11} \cmidrule(lr){12-13} \cmidrule(lr){14-15}
&&&&&&& R@1 & R@5 & R@1 & R@5 & R@1 & R@5 & R@1 & R@5\\
\midrule
& &\ding{51} & & & & &84.9 &87.7 &86.8 &90.1 &71.6 &80.2 &75.6 & 79.5
\\
  \ding{51} &\ding{51} &\ding{51} & & &  & &90.4 &91.6 &91.5 &92.9 & 75.9 & 88.1 &80.4 &84.5 \\
  & & &\ding{51}&\ding{51}& & &91.9 & 93.1 & 92.1 &93.8 & 76.4 & 88.7 &82.5 &87.7 \\
 & &\ding{51}&\ding{51}&\ding{51}& & & 93.7 &95.5 &95.5 & 97.6 &80.3 &89.8 &86.4 &91.2 \\
 \ding{51}&\ding{51}&\ding{51}&\ding{51}& & \ding{51}  & & 92.2 & 94.5 & 94.3 & 95.4 & 78.8 & 89.4 & 85.4 & 89.8 \\
 \ding{51}&\ding{51}&\ding{51}&\ding{51}&\ding{51}& &\ding{51} &93.9 & 95.9 & 96.1 & 97.4 &78.3 & 89.3 & 85.8 & 90.3 \\
 
\ding{51}&\ding{51}&\ding{51}&\ding{51}&\ding{51}& & &94.1  & 96.2  & 96.3 & 98.9 & 81.4 & 91.5 & 87.4 & 92.5 \\
\bottomrule
\end{tabular}}}
\caption{\textbf{Ablation studies} on 2D–3D token selection, aggregation strategies, and 3D pose injection, and analyze their impact on the final results on additional datasets.}
\label{tab:ablation}
\vspace{-0.7cm}
\end{table*}

\begin{figure*}[h]
    \centering
    \includegraphics[width=\linewidth]{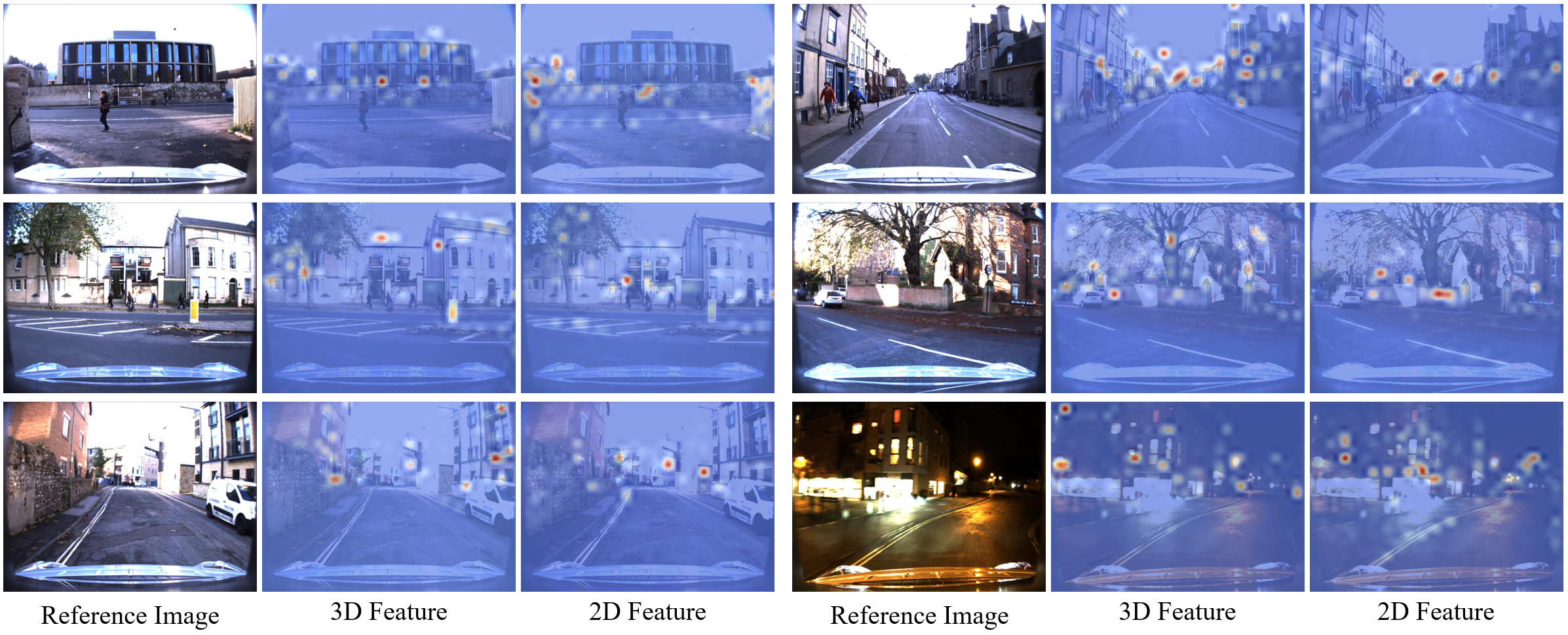}
    \caption{\textbf{Activation heatmaps of 3D and 2D features} in the Oxford dataset~\cite{oxfordrobo}. 
The left column shows reference images, and the two right columns their corresponding 3D and 2D feature heatmaps. 
Note how the different feature types fire at different locations in the image, illustrating their complementary nature.}
    \label{fig:feature}
    \vspace{-0.4cm}
\end{figure*}

 \begin{table}[]
 \centering
\scalebox{0.9}{
\setlength{\tabcolsep}{1.5mm}{
\begin{tabular}{llll}
\toprule
\multirow{2}{*}{\#frames at test time} & \multicolumn{3}{c}{Oxford2} \\ \cline{2-4} 
                         & R@1    & R@5    & R@10   \\ \midrule
3       & 70.3       &  87.2      &  91.7      \\
5 (training setup)       & 80.6       &  90.3      &   93.9     \\
10       & 89.1       & 93.1       & 95.3       \\
15      &  \textbf{92.4}      &  \textbf{94.9}      &  \textbf{96.6}      \\ \bottomrule
\end{tabular}}}
\caption{\textbf{Results with varying sequence length at test time}. UniVPR-3D generalizes for sequence lengths 
differing from the training setting, maintaining strong performance.}
\label{tab:seqlength}
\vspace{-0.8cm}
\end{table}

\subsection{Ablation Studies}

\noindent \textbf{Effect of the different descriptors.}
Our final descriptor consists of both a texture descriptor and a 3D geometric descriptor. 
Each descriptor further integrates five distinct types of tokens. 
To validate the contribution of each token, we first conduct ablation experiments 
on the 2D and 3D patch descriptors, shown in Tab.~\ref{tab:ablation}. 
The results indicate that both 2D and 3D patch tokens have a significant contribution to the final performance. 
Specifically, the 2D patch tokens capture rich texture details of the environment, 
which are particularly beneficial in scenes with sparse structural cues, 
while the 3D patch tokens encode geometric structures, 
offering stronger robustness under illumination and weather variations. 
In addition, the 2D \texttt{cls} and register tokens provide semantic and category-level information 
that produces an additional slight improvement of the recall.

\noindent \textbf{Effect of 3D Pose Injection.} Beyond token selection and aggregation strategies, we also investigated the explicit injection of 3D pose information. Specifically, we experimented with a pose projector using an MLP with Fourier position embeddings to inject relative pose information for sequence matching. However, as shown in Tab.~\ref{tab:ablation}, we observed that this design does not improve retrieval performance and even slightly degrades the results. This suggests that our geometry-aware 3D tokens already implicitly capture sufficient spatial relationships, making explicit pose injection redundant.

To further analyze the complementary roles of 2D and 3D tokens, 
we visualize their relative importance in Fig.~\ref{fig:feature}. 
The figure illustrates the weights of tokens not assigned to the “dustbin,” 
revealing the regions that the network deems informative. We can see that the network effectively discards uninformative regions such as skies, roads, and dynamic objects.
Moreover, when comparing the two heatmaps, we observe that 2D features primarily focus on texture-rich areas, 
such as posters, kiosks, and bicycles, whereas 3D features attend more to geometric and structural elements 
like walls and buildings, reflecting a stronger spatial understanding.

\noindent\textbf{Effect of sequence length.}
We evaluate the impact of the sequence length to assess the adaptability of our multi-frame feature aggregation module to varying input sizes. 
The results are presented in Tab.~\ref{tab:seqlength}. 
Although UniPR-3D is trained only with sequences of length 5, 
it generalizes effectively to sequences of different lengths at test time, 
consistently achieving strong performance across all settings. 
Moreover, recall consistently improves as more images are aggregated, even beyond the original training length, showing that our aggregation design scales robustly to arbitrary numbers of frames.

\vspace{-0.3cm}
\subsection{Conclusion}
In this paper, we propose UniPR-3D, the first 3D token-based VPR method that supports universal scene recognition and outperforms previous baselines by a substantial margin. Moreover, our framework is capable of handling both frame-to-frame and sequence-level retrievals. We design 2D and 3D feature aggregation strategies tailored to the characteristics of different types of 2D and 3D tokens. For sequence-level retrieval, we further propose a multi-frame feature aggregation method for variable-length sequence matching. The final descriptor is constructed by combining the 2D and 3D descriptors. Our extensive experiments demonstrate the effectiveness of the proposed method, achieving strong performance in both single frame and sequence matching. This work provides new insights for the VPR community, 
highlighting the potential of transitioning from conventional 
2D token–based descriptors to 3D token–based representations.

\textbf{Acknowledgement}
This work was supported by National Key R\&D Program of China (Grant No.2024YFB4708900). It was also supported in part by the Natural Science Foundation of
China under Grant 62225309,U24A20278, 62361166632.


\bibliographystyle{splncs04}
\bibliography{main}

@String(CVPR= {IEEE Conf. Comput. Vis. Pattern Recog.})

@String(ICLR = {Int. Conf. Learn. Represent.})

@String(AAAI = {AAAI})

@String(CVPR  = {CVPR})

@String(ICLR  = {ICLR})

@article{galvez2012bags,
  title={Bags of binary words for fast place recognition in image sequences},
  author={G{\'a}lvez-L{\'o}pez, Dorian and Tardos, Juan D},
  journal={IEEE Transactions on robotics},
  volume={28},
  number={5},
  pages={1188--1197},
  year={2012},
  publisher={IEEE}
}

@InProceedings{Sferrazza_2025_CVPR,
    author    = {Sferrazza, Davide and Berton, Gabriele and Trivigno, Gabriele and Masone, Carlo},
    title     = {To Match or Not to Match: Revisiting Image Matching for Reliable Visual Place Recognition},
    booktitle = {Proceedings of the IEEE/CVF Conference on Computer Vision and Pattern Recognition (CVPR) Workshops},
    month     = {June},
    year      = {2025},
    pages     = {2874-2885}
}

@article{orbslam3,
  title={Orb-slam3: An accurate open-source library for visual, visual--inertial, and multimap slam},
  author={Campos, Carlos and Elvira, Richard and Rodr{\'\i}guez, Juan J G{\'o}mez and Montiel, Jos{\'e} MM and Tard{\'o}s, Juan D},
  journal={IEEE Transactions on Robotics},
  volume={37},
  number={6},
  pages={1874--1890},
  year={2021},
  publisher={IEEE}
}

@inproceedings{adam,
  title={Adam: A Method for Stochastic Optimization},
  author={Kingma, Diederik P and Ba, Jimmy},
  booktitle={ICLR},
  year={2015}
}

@article{prosgnerf,
  title={ProSGNeRF: Progressive Dynamic Neural Scene Graph with Frequency Modulated Auto-Encoder in Urban Scenes},
  author={Deng, Tianchen and Liu, Siyang and Wang, Xuan and Liu, Yejia and Wang, Danwei and Chen, Weidong},
  journal={arXiv preprint arXiv:2312.09076},
  year={2023}
}

@inproceedings{plgslam,
  title={Plgslam: Progressive neural scene represenation with local to global bundle adjustment},
  author={Deng, Tianchen and Shen, Guole and Qin, Tong and Wang, Jianyu and Zhao, Wentao and Wang, Jingchuan and Wang, Danwei and Chen, Weidong},
  booktitle={Proceedings of the IEEE/CVF Conference on Computer Vision and Pattern Recognition},
  pages={19657--19666},
  year={2024}
}

@article{oxfordrobo,
  title={1 year, 1000 km: The oxford robotcar dataset},
  author={Maddern, Will and Pascoe, Geoffrey and Linegar, Chris and Newman, Paul},
  journal={The International Journal of Robotics Research},
  volume={36},
  number={1},
  pages={3--15},
  year={2017},
  publisher={SAGE Publications Sage UK: London, England}
}

@ARTICLE{neslam,
  author={Deng, Tianchen and Wang, Yanbo and Xie, Hongle and Wang, Hesheng and Guo, Rui and Wang, Jingchuan and Wang, Danwei and Chen, Weidong},
  journal={IEEE Transactions on Automation Science and Engineering}, 
  title={NeSLAM: Neural Implicit Mapping and Self-Supervised Feature Tracking With Depth Completion and Denoising}, 
  year={2025},
  volume={},
  number={},
  pages={1-1},
  doi={10.1109/TASE.2025.3541064}}

@article{deng2025mcnslam,
      title={MCN-SLAM: Multi-Agent Collaborative Neural SLAM with Hybrid Implicit Neural Scene Representation}, 
      author={Tianchen Deng and Guole Shen and Xun Chen and Shenghai Yuan and Hongming Shen and Guohao Peng and Zhenyu Wu and Jingchuan Wang and Lihua Xie and Danwei Wang and Hesheng Wang and Weidong Chen},
      journal={arXiv preprint arXiv:2506.18678},
      year={2025},
}

@article{grsslam,
  title={GRS-SLAM3R: Real-Time Dense SLAM with Gated Recurrent State},
  author={Shen, Guole and Deng, Tianchen and Wang, Yanbo and Chen, Yongtao and Shen, Yilin and Liu, Jiuming and Wang, Jingchuan},
  journal={arXiv preprint arXiv:2509.23737},
  year={2025}
}

@article{lenav,
  title={Learning to Tune Like an Expert: Interpretable and Scene-Aware Navigation via MLLM Reasoning and CVAE-Based Adaptation},
  author={Wang, Yanbo and Fang, Zipeng and Zhao, Lei and Chen, Weidong},
  journal={arXiv preprint arXiv:2507.11001},
  year={2025}
}

@article{mgslam,
  title={MG-SLAM: Structure Gaussian Splatting SLAM with Manhattan World Hypothesis},
  author={Liu, Shuhong and Deng, Tianchen and Zhou, Heng and Li, Liuzhuozheng and Wang, Hongyu and Wang, Danwei and Li, Mingrui},
  journal={IEEE Transactions on Automation Science and Engineering},
  year={2025},
  publisher={IEEE}
}

@inproceedings{mneslam,
  title={Mne-slam: Multi-agent neural slam for mobile robots},
  author={Deng, Tianchen and Shen, Guole and Xun, Chen and Yuan, Shenghai and Jin, Tongxin and Shen, Hongming and Wang, Yanbo and Wang, Jingchuan and Wang, Hesheng and Wang, Danwei and others},
  booktitle={Proceedings of the Computer Vision and Pattern Recognition Conference},
  pages={1485--1494},
  year={2025}
}

@inproceedings{vggt,
  title={Vggt: Visual geometry grounded transformer},
  author={Wang, Jianyuan and Chen, Minghao and Karaev, Nikita and Vedaldi, Andrea and Rupprecht, Christian and Novotny, David},
  booktitle={Proceedings of the Computer Vision and Pattern Recognition Conference},
  pages={5294--5306},
  year={2025}
}

@inproceedings{patchvlad,
  title={Patch-netvlad: Multi-scale fusion of locally-global descriptors for place recognition},
  author={Hausler, Stephen and Garg, Sourav and Xu, Ming and Milford, Michael and Fischer, Tobias},
  booktitle={Proceedings of the IEEE/CVF conference on computer vision and pattern recognition},
  pages={14141--14152},
  year={2021}
}

@inproceedings{netvlad,
  title={NetVLAD: CNN architecture for weakly supervised place recognition},
  author={Arandjelovic, Relja and Gronat, Petr and Torii, Akihiko and Pajdla, Tomas and Sivic, Josef},
  booktitle={Proceedings of the IEEE conference on computer vision and pattern recognition},
  pages={5297--5307},
  year={2016}
}

@inproceedings{transvpr,
  title={Transvpr: Transformer-based place recognition with multi-level attention aggregation},
  author={Wang, Ruotong and Shen, Yanqing and Zuo, Weiliang and Zhou, Sanping and Zheng, Nanning},
  booktitle={Proceedings of the IEEE/CVF Conference on Computer Vision and Pattern Recognition},
  pages={13648--13657},
  year={2022}
}

@inproceedings{r2former,
  title={R2former: Unified retrieval and reranking transformer for place recognition},
  author={Zhu, Sijie and Yang, Linjie and Chen, Chen and Shah, Mubarak and Shen, Xiaohui and Wang, Heng},
  booktitle={Proceedings of the IEEE/CVF Conference on Computer Vision and Pattern Recognition},
  pages={19370--19380},
  year={2023}
}

@inproceedings{vlad,
  title={Aggregating local descriptors into a compact image representation},
  author={J{\'e}gou, Herv{\'e} and Douze, Matthijs and Schmid, Cordelia and P{\'e}rez, Patrick},
  booktitle={2010 IEEE computer society conference on computer vision and pattern recognition},
  pages={3304--3311},
  year={2010},
  organization={IEEE}
}

@article{anyloc,
  title={Anyloc: Towards universal visual place recognition},
  author={Keetha, Nikhil and Mishra, Avneesh and Karhade, Jay and Jatavallabhula, Krishna Murthy and Scherer, Sebastian and Krishna, Madhava and Garg, Sourav},
  journal={IEEE Robotics and Automation Letters},
  volume={9},
  number={2},
  pages={1286--1293},
  year={2023},
  publisher={IEEE}
}

@inproceedings{salad,
  title={Optimal transport aggregation for visual place recognition},
  author={Izquierdo, Sergio and Civera, Javier},
  booktitle={Proceedings of the ieee/cvf conference on computer vision and pattern recognition},
  pages={17658--17668},
  year={2024}
}

@article{dinov2,
  title={Dinov2: Learning robust visual features without supervision},
  author={Oquab, Maxime and Darcet, Timoth{\'e}e and Moutakanni, Th{\'e}o and Vo, Huy and Szafraniec, Marc and Khalidov, Vasil and Fernandez, Pierre and Haziza, Daniel and Massa, Francisco and El-Nouby, Alaaeldin and others},
  journal={arXiv preprint arXiv:2304.07193},
  year={2023}
}

@article{gem,
  title={Fine-tuning CNN image retrieval with no human annotation},
  author={Radenovi{\'c}, Filip and Tolias, Giorgos and Chum, Ond{\v{r}}ej},
  journal={IEEE transactions on pattern analysis and machine intelligence},
  volume={41},
  number={7},
  pages={1655--1668},
  year={2018},
  publisher={IEEE}
}

@inproceedings{mixvpr,
  title={Mixvpr: Feature mixing for visual place recognition},
  author={Ali-Bey, Amar and Chaib-Draa, Brahim and Giguere, Philippe},
  booktitle={Proceedings of the IEEE/CVF winter conference on applications of computer vision},
  pages={2998--3007},
  year={2023}
}

@inproceedings{ot,
  title={A survey of optimal transport for computer graphics and computer vision},
  author={Bonneel, Nicolas and Digne, Julie},
  booktitle={Computer Graphics Forum},
  volume={42},
  number={2},
  pages={439--460},
  year={2023},
  organization={Wiley Online Library}
}

@inproceedings{seqslam,
  title={SeqSLAM: Visual route-based navigation for sunny summer days and stormy winter nights},
  author={Milford, Michael J and Wyeth, Gordon F},
  booktitle={2012 IEEE international conference on robotics and automation},
  pages={1643--1649},
  year={2012},
  organization={IEEE}
}

@inproceedings{fast,
  title={Fast and Memory Efficient Graph Optimization via ICM for Visual Place Recognition.},
  author={Schubert, Stefan and Neubert, Peer and Protzel, Peter},
  booktitle={Robotics: Science and Systems},
  volume={2},
  year={2021}
}

@article{longslam,
  title={Long-term visual simultaneous localization and mapping: Using a bayesian persistence filter-based global map prediction},
  author={Deng, Tianchen and Xie, Hongle and Wang, Jingchuan and Chen, Weidong},
  journal={IEEE Robotics \& Automation Magazine},
  volume={30},
  number={1},
  pages={36--49},
  year={2023},
  publisher={IEEE}
}

@inproceedings{seqmatchnet,
  title={Seqmatchnet: Contrastive learning with sequence matching for place recognition \& relocalization},
  author={Garg, Sourav and Vankadari, Madhu and Milford, Michael},
  booktitle={Conference on Robot Learning},
  pages={429--443},
  year={2022},
  organization={PMLR}
}

@article{seqvlad,
  title={Learning sequential descriptors for sequence-based visual place recognition},
  author={Mereu, Riccardo and Trivigno, Gabriele and Berton, Gabriele and Masone, Carlo and Caputo, Barbara},
  journal={IEEE Robotics and Automation Letters},
  volume={7},
  number={4},
  pages={10383--10390},
  year={2022},
  publisher={IEEE}
}

@ARTICLE{svpr,
  author={Zhao, Junqiao and Zhang, Fenglin and Cai, Yingfeng and Tian, Gengxuan and Mu, Wenjie and Ye, Chen and Feng, Tiantian},
  journal={IEEE Robotics and Automation Letters}, 
  title={Learning Sequence Descriptor Based on Spatio-Temporal Attention for Visual Place Recognition}, 
  year={2024},
  volume={9},
  number={3},
  pages={2351-2358},
  doi={10.1109/LRA.2024.3354627}}

@article{seqnet,
  title={Seqnet: Learning descriptors for sequence-based hierarchical place recognition},
  author={Garg, Sourav and Milford, Michael},
  journal={IEEE Robotics and Automation Letters},
  volume={6},
  number={3},
  pages={4305--4312},
  year={2021},
  publisher={IEEE}
}

@inproceedings{long-match,
  title={Towards life-long visual localization using an efficient matching of binary sequences from images},
  author={Arroyo, Roberto and Alcantarilla, Pablo F and Bergasa, Luis M and Romera, Eduardo},
  booktitle={2015 IEEE international conference on robotics and automation (ICRA)},
  pages={6328--6335},
  year={2015},
  organization={IEEE}
}

@article{bow,
  title={Bags of binary words for fast place recognition in image sequences},
  author={G{\'a}lvez-L{\'o}pez, Dorian and Tardos, Juan D},
  journal={IEEE Transactions on robotics},
  volume={28},
  number={5},
  pages={1188--1197},
  year={2012},
  publisher={IEEE}
}

@article{globallocalization,
  title={Localization in urban environments using a panoramic gist descriptor},
  author={Murillo, Ana C and Singh, Gautam and Kosecka, Jana and Guerrero, Jos{\'e} Jes{\'u}s},
  journal={IEEE Transactions on Robotics},
  volume={29},
  number={1},
  pages={146--160},
  year={2012},
  publisher={IEEE}
}

@inproceedings{brief,
  title={Brief-gist-closing the loop by simple means},
  author={S{\"u}nderhauf, Niko and Protzel, Peter},
  booktitle={2011 IEEE/RSJ International Conference on Intelligent Robots and Systems},
  pages={1234--1241},
  year={2011},
  organization={IEEE}
}

@article{gsv,
  title={Gsv-cities: Toward appropriate supervised visual place recognition},
  author={Ali-bey, Amar and Chaib-draa, Brahim and Giguere, Philippe},
  journal={Neurocomputing},
  volume={513},
  pages={194--203},
  year={2022},
  publisher={Elsevier}
}

@inproceedings{superglue,
  title={Superglue: Learning feature matching with graph neural networks},
  author={Sarlin, Paul-Edouard and DeTone, Daniel and Malisiewicz, Tomasz and Rabinovich, Andrew},
  booktitle={Proceedings of the IEEE/CVF conference on computer vision and pattern recognition},
  pages={4938--4947},
  year={2020}
}

@article{sinkhorn,
  title={Sinkhorn distances: Lightspeed computation of optimal transport},
  author={Cuturi, Marco},
  journal={Advances in neural information processing systems},
  volume={26},
  year={2013}
}

@inproceedings{cosplace,
  title={Rethinking visual geo-localization for large-scale applications},
  author={Berton, Gabriele and Masone, Carlo and Caputo, Barbara},
  booktitle={Proceedings of the IEEE/CVF Conference on Computer Vision and Pattern Recognition},
  pages={4878--4888},
  year={2022}
}

@inproceedings{msls,
  title={Mapillary street-level sequences: A dataset for lifelong place recognition},
  author={Warburg, Frederik and Hauberg, Soren and Lopez-Antequera, Manuel and Gargallo, Pau and Kuang, Yubin and Civera, Javier},
  booktitle={Proceedings of the IEEE/CVF conference on computer vision and pattern recognition},
  pages={2626--2635},
  year={2020}
}

@inproceedings{multiloss,
  title={Multi-similarity loss with general pair weighting for deep metric learning},
  author={Wang, Xun and Han, Xintong and Huang, Weilin and Dong, Dengke and Scott, Matthew R},
  booktitle={Proceedings of the IEEE/CVF conference on computer vision and pattern recognition},
  pages={5022--5030},
  year={2019}
}

@inproceedings{eigenplaces,
  title={Eigenplaces: Training viewpoint robust models for visual place recognition},
  author={Berton, Gabriele and Trivigno, Gabriele and Caputo, Barbara and Masone, Carlo},
  booktitle={Proceedings of the IEEE/CVF International Conference on Computer Vision},
  pages={11080--11090},
  year={2023}
}

@inproceedings{Pittsburg,
  title={Visual place recognition with repetitive structures},
  author={Torii, Akihiko and Sivic, Josef and Pajdla, Tomas and Okutomi, Masatoshi},
  booktitle={Proceedings of the IEEE conference on computer vision and pattern recognition},
  pages={883--890},
  year={2013}
}

@inproceedings{nordland,
  title={Are we there yet? Challenging SeqSLAM on a 3000 km journey across all four seasons},
  author={S{\"u}nderhauf, Niko and Neubert, Peer and Protzel, Peter},
  booktitle={Proc. of workshop on long-term autonomy, IEEE international conference on robotics and automation (ICRA)},
  pages={2013},
  year={2013},
  organization={Citeseer}
}

@article{flashattention,
  title={Flashattention-2: Faster attention with better parallelism and work partitioning},
  author={Dao, Tri},
  journal={arXiv preprint arXiv:2307.08691},
  year={2023}
}

@article{delta,
  title={Delta descriptors: Change-based place representation for robust visual localization},
  author={Garg, Sourav and Harwood, Ben and Anand, Gaurangi and Milford, Michael},
  journal={IEEE Robotics and Automation Letters},
  volume={5},
  number={4},
  pages={5120--5127},
  year={2020},
  publisher={IEEE}
}

@article{sped,
  title={Learning context flexible attention model for long-term visual place recognition},
  author={Chen, Zetao and Liu, Lingqiao and Sa, Inkyu and Ge, Zongyuan and Chli, Margarita},
  journal={IEEE Robotics and Automation Letters},
  volume={3},
  number={4},
  pages={4015--4022},
  year={2018},
  publisher={IEEE}
}

@article{sequencematch,
  title={Improving Visual Place Recognition with Sequence-Matching Receptiveness Prediction},
  author={Hussaini, Somayeh and Fischer, Tobias and Milford, Michael},
  journal={arXiv preprint arXiv:2503.06840},
  year={2025}
}

@article{casevpr,
  title={CaseVPR: Correlation-Aware Sequential Embedding for Sequence-to-Frame Visual Place Recognition},
  author={Li, Heshan and Peng, Guohao and Zhang, Jun and Wen, Mingxing and Ma, Yingchong and Wang, Danwei},
  journal={IEEE Robotics and Automation Letters},
  year={2025},
  publisher={IEEE}
}

@article{jist,
  title={Jist: Joint image and sequence training for sequential visual place recognition},
  author={Berton, Gabriele and Trivigno, Gabriele and Caputo, Barbara and Masone, Carlo},
  journal={IEEE Robotics and Automation Letters},
  volume={9},
  number={2},
  pages={1310--1317},
  year={2023},
  publisher={IEEE}
}

@inproceedings{sarlin2019coarse,
  title={From coarse to fine: Robust hierarchical localization at large scale},
  author={Sarlin, Paul-Edouard and Cadena, Cesar and Siegwart, Roland and Dymczyk, Marcin},
  booktitle={Proceedings of the IEEE/CVF conference on computer vision and pattern recognition},
  pages={12716--12725},
  year={2019}
}

@inproceedings{taira2018inloc,
  title={InLoc: Indoor visual localization with dense matching and view synthesis},
  author={Taira, Hajime and Okutomi, Masatoshi and Sattler, Torsten and Cimpoi, Mircea and Pollefeys, Marc and Sivic, Josef and Pajdla, Tomas and Torii, Akihiko},
  booktitle={Proceedings of the IEEE conference on computer vision and pattern recognition},
  pages={7199--7209},
  year={2018}
}

@inproceedings{zhou2024nerfect,
  title={The nerfect match: Exploring nerf features for visual localization},
  author={Zhou, Qunjie and Maximov, Maxim and Litany, Or and Leal-Taix{\'e}, Laura},
  booktitle={European Conference on Computer Vision},
  pages={108--127},
  year={2024},
  organization={Springer}
}

@inproceedings{izquierdo2024close,
  title={Close, but not there: Boosting geographic distance sensitivity in visual place recognition},
  author={Izquierdo, Sergio and Civera, Javier},
  booktitle={European Conference on Computer Vision},
  pages={240--257},
  year={2024},
  organization={Springer}
}

@inproceedings{berton2025megaloc,
  title={Megaloc: One retrieval to place them all},
  author={Berton, Gabriele and Masone, Carlo},
  booktitle={Proceedings of the Computer Vision and Pattern Recognition Conference Workshops},
  pages={2861--2867},
  year={2025}
}

@article{Schubert2023vpr,
  title={{Visual Place Recognition: A Tutorial}},
  author={Schubert, Stefan and Neubert, Peer and Garg, Sourav and Milford, Michael and Fischer, Tobias},
  journal={IEEE Robotics \& Automation Magazine},
  year={2023},
  publisher={IEEE}
}

@article{garg2021your,
  title={Where is your place, visual place recognition?},
  author={Garg, Sourav and Fischer, Tobias and Milford, Michael},
  journal={arXiv preprint arXiv:2103.06443},
  year={2021}
}

@article{lowry2015visual,
  title={Visual place recognition: A survey},
  author={Lowry, Stephanie and S{\"u}nderhauf, Niko and Newman, Paul and Leonard, John J and Cox, David and Corke, Peter and Milford, Michael J},
  journal={IEEE Transactions on Robotics},
  volume={32},
  number={1},
  pages={1--19},
  year={2015},
  publisher={IEEE}
}

@article{masone2021survey,
  title={A survey on deep visual place recognition},
  author={Masone, Carlo and Caputo, Barbara},
  journal={IEEE Access},
  volume={9},
  pages={19516--19547},
  year={2021},
  publisher={IEEE}
}

@article{zhang2021visual,
  title={Visual place recognition: A survey from deep learning perspective},
  author={Zhang, Xiwu and Wang, Lei and Su, Yan},
  journal={Pattern Recognition},
  volume={113},
  pages={107760},
  year={2021},
  publisher={Elsevier}
}

@article{milford2025going,
  title={Going Places: Place Recognition in Artificial and Natural Systems},
  author={Milford, Michael and Fischer, Tobias},
  journal={Annual Review of Control, Robotics, and Autonomous Systems},
  volume={9},
  year={2025},
  publisher={Annual Reviews}
}

@article{facil2019condition,
  title={Condition-invariant multi-view place recognition},
  author={Facil, Jose M and Olid, Daniel and Montesano, Luis and Civera, Javier},
  journal={arXiv preprint arXiv:1902.09516},
  year={2019}
}

@inproceedings{sivic2003video,
  title={Video Google: A text retrieval approach to object matching in videos},
  author={Sivic and Zisserman},
  booktitle={Proceedings ninth IEEE international conference on computer vision},
  pages={1470--1477},
  year={2003},
  organization={IEEE}
}

@inproceedings{sfpnet,
  title={Sfpnet: Sparse focal point network for semantic segmentation on general lidar point clouds},
  author={Wang, Yanbo and Zhao, Wentao and Cao, Chuan and Deng, Tianchen and Wang, Jingchuan and Chen, Weidong},
  booktitle={European Conference on Computer Vision},
  pages={403--421},
  year={2024},
  organization={Springer}
}

@article{lv1,
  title={Neural network-based nonconservative predefined-time backstepping control for uncertain strict-feedback nonlinear systems},
  author={Lv, Jixing and Ju, Xiaozhe and Wang, Changhong},
  journal={IEEE Transactions on Neural Networks and Learning Systems},
  year={2023},
  publisher={IEEE}
}

@article{lv2,
  title={Adaptive distributed observer design for nonlinear multiagent systems},
  author={Lv, Jixing and Wang, Changhong and Xie, Lihua},
  journal={Automatica},
  volume={183},
  pages={112625},
  year={2026},
  publisher={Elsevier}
}

@ARTICLE{yang1,
  author={Yang, Changzhi and Pan, Huihui and Wang, Jue},
  journal={IEEE Transactions on Vehicular Technology}, 
  title={STGCNFormer: Spatio-Temporal Dual-Stream Graph Convolutional Networks and Transformers for Traffic Forecasting}, 
  year={2025},
  volume={74},
  number={10},
  pages={15254-15263},
  doi={10.1109/TVT.2025.3572622}}

@ARTICLE{yang2,
  author={Yang, Changzhi and Pan, Huihui and Wang, Jue and Hong, Yuanduo},
  journal={IEEE Transactions on Image Processing}, 
  title={TrajDiff: Trajectory Prediction with Diffusion Probabilistic Models}, 
  year={2025},
  volume={},
  number={},
  pages={1-14},
  doi={10.1109/TIP.2025.3640001}}

@ARTICLE{10668846,
  author={Liu, Qiming and Xin, Haoran and Liu, Zhe and Wang, Hesheng},
  journal={IEEE Transactions on Pattern Analysis and Machine Intelligence}, 
  title={Integrating Neural Radiance Fields End-to-End for Cognitive Visuomotor Navigation}, 
  year={2024},
  volume={46},
  number={12},
  pages={11200-11215},
  doi={10.1109/TPAMI.2024.3455252}}

@ARTICLE{10486967,
  author={Liu, Qiming and Chen, Nanxi and Liu, Zhe and Wang, Hesheng},
  journal={IEEE Transactions on Industrial Informatics}, 
  title={Toward Learning-Based Visuomotor Navigation With Neural Radiance Fields}, 
  year={2024},
  volume={20},
  number={6},
  pages={8907-8916},
  doi={10.1109/TII.2024.3378829}}

@article{unilgl,
  title={UniLGL: Learning Uniform Place Recognition for FOV-limited/Panoramic LiDAR Global Localization},
  author={Shen, Hongming and Chen, Xun and Hui, Yulin and Wu, Zhenyu and Wang, Wei and Lyu, Qiyang and Deng, Tianchen and Wang, Danwei},
  journal={arXiv preprint arXiv:2507.12194},
  year={2025}
}

@article{3dscenerepresentation,
      title={What Is The Best 3D Scene Representation for Robotics? From Geometric to Foundation Models}, 
      author={Tianchen Deng and Yue Pan and Shenghai Yuan and Dong Li and Chen Wang and Mingrui Li and Long Chen and Lihua Xie and Danwei Wang and Jingchuan Wang and Javier Civera and Hesheng Wang and Weidong Chen},
      year={2025},
      journal={arXiv preprint arXiv:2512.03422}, 
}

@inproceedings{apgem,
  title={Learning with average precision: Training image retrieval with a listwise loss},
  author={Revaud, Jerome and Almaz{\'a}n, Jon and Rezende, Rafael S and Souza, Cesar Roberto de},
  booktitle={Proceedings of the IEEE/CVF international conference on computer vision},
  pages={5107--5116},
  year={2019}
}

@inproceedings{lu2024cricavpr,
  title={Cricavpr: Cross-image correlation-aware representation learning for visual place recognition},
  author={Lu, Feng and Lan, Xiangyuan and Zhang, Lijun and Jiang, Dongmei and Wang, Yaowei and Yuan, Chun},
  booktitle={Proceedings of the IEEE/CVF conference on computer vision and pattern recognition},
  pages={16772--16782},
  year={2024}
}

@inproceedings{ali2024boq,
  title={Boq: A place is worth a bag of learnable queries},
  author={Ali-Bey, Amar and Chaib-Draa, Brahim and Giguere, Philippe},
  booktitle={Proceedings of the IEEE/CVF conference on computer vision and pattern recognition},
  pages={17794--17803},
  year={2024}
}

@inproceedings{tokyo247,
  title={24/7 place recognition by view synthesis},
  author={Torii, Akihiko and Arandjelovic, Relja and Sivic, Josef and Okutomi, Masatoshi and Pajdla, Tomas},
  booktitle={Proceedings of the IEEE conference on computer vision and pattern recognition},
  pages={1808--1817},
  year={2015}
}

@inproceedings{sfxl,
  title={Rethinking visual geo-localization for large-scale applications},
  author={Berton, Gabriele and Masone, Carlo and Caputo, Barbara},
  booktitle={Proceedings of the IEEE/CVF Conference on Computer Vision and Pattern Recognition},
  pages={4878--4888},
  year={2022}
}

@inproceedings{sfxl_extraqueries,
  title={Are local features all you need for cross-domain visual place recognition?},
  author={Barbarani, Giovanni and Mostafa, Mohamad and Bayramov, Hajali and Trivigno, Gabriele and Berton, Gabriele and Masone, Carlo and Caputo, Barbara},
  booktitle={Proceedings of the IEEE/CVF Conference on Computer Vision and Pattern Recognition},
  pages={6155--6165},
  year={2023}
}

@ARTICLE{selavprpp,
author={Lu, Feng and Jin, Tong and Lan, Xiangyuan and Zhang, Lijun and Liu, Yunpeng and Wang, Yaowei and Yuan, Chun}, 
  title={SelaVPR++: Towards Seamless Adaptation of Foundation Models for Efficient Place Recognition},
  journal={IEEE Transactions on Pattern Analysis and Machine Intelligence},
  year={2026},
  volume={48},
  number={3},
  pages={2731-2748}}

@inproceedings{selavpr,
  title={Towards Seamless Adaptation of Pre-trained Models for Visual Place Recognition},
  author={Lu, Feng and Zhang, Lijun and Lan, Xiangyuan and Dong, Shuting and Wang, Yaowei and Yuan, Chun},
  booktitle={The Twelfth International Conference on Learning Representations},
  year={2024}
}

@article{ma2025controllable,
  title={Controllable Video Generation: A Survey},
  author={Ma, Yue and Feng, Kunyu and Hu, Zhongyuan and Wang, Xinyu and Wang, Yucheng and Zheng, Mingzhe and He, Xuanhua and Zhu, Chenyang and Liu, Hongyu and He, Yingqing and others},
  journal={arXiv preprint arXiv:2507.16869},
  year={2025}
}

@article{ma2025followcreation,
  title={Follow-Your-Creation: Empowering 4D Creation through Video Inpainting},
  author={Ma, Yue and Feng, Kunyu and Zhang, Xinhua and Liu, Hongyu and Zhang, David Junhao and Xing, Jinbo and Zhang, Yinhan and Yang, Ayden and Wang, Zeyu and Chen, Qifeng},
  journal={arXiv preprint arXiv:2506.04590},
  year={2025}
}

@inproceedings{ma2024followpose,
  title={Follow your pose: Pose-guided text-to-video generation using pose-free videos},
  author={Ma, Yue and He, Yingqing and Cun, Xiaodong and Wang, Xintao and Chen, Siran and Li, Xiu and Chen, Qifeng},
  booktitle={Proceedings of the AAAI Conference on Artificial Intelligence},
  volume={38},
  number={5},
  pages={4117--4125},
  year={2024}
}

@inproceedings{chengmoca,
  title={MoCa: Modeling Object Consistency for 3D Camera Control in Video Generation},
  author={Cheng, Zhijing and Zhang, Xuancheng and Di, Donglin and Wei, Chen and Li, Hao and Yang, Xun},
  booktitle={The Fourteenth International Conference on Learning Representations}
}

@article{tosi2026nerfs,
  title={How nerfs and 3d gaussian splatting are reshaping slam: a survey},
  author={Tosi, Fabio and Zhang, Youmin and Gong, Ziren and Mattoccia, Stefano and Oswald, Martin R and Sandstrom, Erik and Poggi, Matteo},
  journal={IEEE Transactions on Robotics},
  year={2026},
  publisher={IEEE}
}

@article{gong2025dino,
  title={DINO-SLAM: DINO-informed RGB-D SLAM for Neural Implicit and Explicit Representations},
  author={Gong, Ziren and Li, Xiaohan and Tosi, Fabio and Zhang, Youmin and Mattoccia, Stefano and Wu, Jun and Poggi, Matteo},
  journal={arXiv preprint arXiv:2507.19474},
  year={2025}
}

@article{song2025accelerating,
  title={Accelerating vision-language-action model integrated with action chunking via parallel decoding},
  author={Song, Wenxuan and Chen, Jiayi and Ding, Pengxiang and Zhao, Han and Zhao, Wei and Zhong, Zhide and Ge, Zongyuan and Ma, Jun and Li, Haoang},
  journal={arXiv preprint arXiv:2503.02310},
  year={2025}
}

@inproceedings{song2026reconvla,
  title={Reconvla: Reconstructive vision-language-action model as effective robot perceiver},
  author={Song, Wenxuan and Zhou, Ziyang and Zhao, Han and Chen, Jiayi and Ding, Pengxiang and Yan, Haodong and Huang, Yuxin and Tang, Feilong and Wang, Donglin and Li, Haoang},
  booktitle={Proceedings of the AAAI Conference on Artificial Intelligence},
  volume={40},
  number={22},
  pages={18549--18557},
  year={2026}
}

@article{dens3r,
      title={Dens3R: A Foundation Model for 3D Geometry Prediction}, 
      author={Xianze Fang and Jingnan Gao and Zhe Wang and Zhuo Chen and Xingyu Ren and Jiangjing Lyu and Qiaomu Ren and Zhonglei Yang and Xiaokang Yang and Yichao Yan and Chengfei Lyu},
      journal={arXiv preprint arXiv:2507.16290},
      year={2025}
}

@article{MoRE2025,
  title={MoRE: 3D Visual Geometry Reconstruction Meets Mixture-of-Experts}, 
  author={Jingnan Gao and Zhe Wang and Xianze Fang and Xingyu Ren and Zhuo Chen and Shengqi Liu and Yuhao Cheng and Jiangjing Lyu and Xiaokang Yang and Yichao Yan},
  journal={arXiv preprint arXiv:2510.27234},
  year={2025}
}

@article{liu2026driveva,
  title={Driveva: Video action models are zero-shot drivers},
  author={Liu, Mengmeng and Zhang, Diankun and Liu, Jiuming and Cui, Jianfeng and Xie, Hongwei and Chen, Guang and Ye, Hangjun and Yang, Michael Ying and Nex, Francesco and Cheng, Hao},
  journal={arXiv preprint arXiv:2604.04198},
  year={2026}
}

@inproceedings{liu2026streamvlo,
  title={StreamVLO: Streaming Visual-LiDAR Odometry with Cumulative Drift Compensation},
  author={Liu, Mengmeng and Liu, Jiuming and Yang, Michael Ying and Jiang, Chaokang and Li, Jiangtao and Zhang, Yunpeng and Wang, Hesheng and Nex, Francesco and Cheng, Hao},
  booktitle={Proceedings of the IEEE/CVF Conference on Computer Vision and Pattern Recognition},
  pages={39086--39097},
  year={2026}
}

@article{tu2025role,
  title={The role of world models in shaping autonomous driving: A comprehensive survey},
  author={Tu, Sifan and Zhou, Xin and Liang, Dingkang and Jiang, Xingyu and Zhang, Yumeng and Li, Xiaofan and Bai, Xiang},
  journal={arXiv preprint arXiv:2502.10498},
  year={2025}
}

@article{wu2026vega,
      title={Generation Models Know Space: Unleashing Implicit 3D Priors for Scene Understanding},
      author={Xianjin Wu and Dingkang Liang and Tianrui Feng and Kui Xia and Yumeng Zhang and Xiaofan Li and Xiao Tan and Xiang Bai},
      journal={arXiv preprint arXiv:2603.19235},
      year={2026}
}

@article{chen2026out,
  title={Out of sight but not out of mind: Hybrid memory for dynamic video world models},
  author={Chen, Kaijin and Liang, Dingkang and Zhou, Xin and Ding, Yikang and Liu, Xiaoqiang and Wan, Pengfei and Bai, Xiang},
  journal={arXiv preprint arXiv:2603.25716},
  year={2026}
}

\end{document}